\documentclass[sigconf]{aamas} 

\usepackage[inline]{enumitem}

\usepackage{multirow}
\usepackage{array}
\usepackage{adjustbox}

\newcommand{\tloA}{TLO$^\text{A}$}

\newcommand{\RP}{R$^\text{P}$}
\newcommand{\RA}{R$^\text{A}$}
\newcommand{\RStar}{R*}

\usepackage{balance} 

\setcopyright{rightsretained}
\acmConference[JAMAAS]{
Autonomous Agents and Multi-Agent Systems: Special Issue on Multi-Objective Decision Making (MODeM)}
{}{Online}{Mannion, Roijers, Vamplew, Dazeley}
\copyrightyear{2021}
\acmYear{2021}
\acmDOI{}
\acmPrice{}
\acmISBN{}

\acmSubmissionID{???}

\title[Improving performance with soft maximin approaches]{Improving performance in multi-objective decision-making in Bottles environments with soft maximin approaches}

\author{Benjamin J. Smith}
\affiliation{
  \department{Center for Translational Neuroscience}
  \institution{University of Oregon}
  \city{Eugene}
  \state{OR}
  }
\email{benjsmith@gmail.com}

\author{Robert Klassert}
\affiliation{
\department{Kirchhoff-Institut f\"ur Physik
}
\institution{Ruprecht-Karls-Universit\"at Heidelberg}
\city{Heidelberg}
\state{Germany}
}
\email{robertklassert@pm.me}

\author{Roland Pihlakas}
\affiliation{
  \department{Independent researcher}
  \institution{Simplify / Macrotec OÜ}
  \city{Tartu, Estonia}}
\email{roland@simplify.ee}

\begin{abstract}

Balancing multiple competing and conflicting objectives is an essential task for any artificial intelligence tasked with satisfying human values or preferences. Conflict arises both from misalignment between individuals with competing values, but also between conflicting value systems held by a single human. Starting with principles of loss-aversion and maximin, we designed a set of soft maximin function approaches to multi-objective decision-making. Bench-marking these functions in a set of previously-developed environments, we found that one new approach in particular, `split-function exp-log loss aversion', learns faster than the thresholded alignment objective method, the state of the art described in \cite{vamplew_potential-based_2021}, on the `BreakableBottles' task. We explore approaches to further improve multi-objective decision-making using soft maximin approaches.

\end{abstract}


\newcommand{\BibTeX}{\rm B\kern-.05em{\sc i\kern-.025em b}\kern-.08em\TeX}

\begin{document}


\pagestyle{fancy}
\fancyhead{}


\maketitle 



\section{Introduction}

A key aim of AI Safety research is to align AI systems to the fulfillment of human preferences \cite{Bostrom2014, russell2019human} or values. There are at least three reasons why this is a multi-objective (MO) problem. First, there are a variety of ethical, legal, and safety-based frameworks \cite{vamplew_human-aligned_2018}, and alignment to any one of these systems is insufficient. Second, even within a specific category--for instance, moral systems--there exist competing accounts of moral outcomes, including amongst philosophers of ethics and morality \cite{bogosian_implementation_2017}. Third, according to the moral intuitionist account of human moral cognition, moral cognition is a plural and contradictory set of social intuitions \cite{haidt2001emotional,sotala2016defining}.

Human values cannot be reliably and consistently reduced to a single outcome or value function in any indisputable way, even at the level of basic biological needs \cite{smith2021multiattributemodel}. Each value is held for its intrinsic, axiomatic worth. When conflicts between fundamental values occur, any possible solution will violate one or more values and is considered unsatisfactory.  

One solution is to design systems that aim for Pareto-optimality, but as the number of objectives increases, it becomes harder to achieve strict Pareto-optimality \cite{rolf_need_2020}. It may then be necessary to look for a heuristic solution that balances Pareto-optimality with the ability to achieve reasonable compromise between objectives. We propose a concave utility function that emphasizes negative rewards more than large positive rewards, without entirely discounting positive rewards. 

It can be argued that having multiple objectives, none of which is allowed to dominate over others, helps to mitigate against Goodhart's law \cite{garrabrant_2017}, "When a measure becomes a target, it ceases to be a good measure" \cite{strathern1997improving}. Goodhart's law manifests when when a pressure is placed upon a particular measure or heuristic is chosen to approximate an ultimate objective that is perhaps hard to directly target; the measure then becomes a de facto objective, often at the expense of achieving the originally intended objective. When the measures are somewhat uncorrelated and domination of any objective is forbidden by a utility aggregation function then particular measures are avoided from bearing too much pressure.

\subsection{Current approaches}

\subsubsection{Multi-objective decision-making in reinforcement learning}
The inclusion of multiple objectives in reinforcement learning tasks was previously explored \cite{vamplew_human-aligned_2018,vamplew_potential-based_2021}  in the form of  \textit{maximin} approaches and \textit{leximin} approaches.
A maximin approach aims to maximize the value of the lowest member of a set--for instance, the outcomes for the least-well-off person in a group of people \cite{rawls2001justice}, or in a multi-objective optimization problem, the outcomes in terms of the objective with the lowest value. A maximin approach may also maximize the value of the least-optimized value (`objective' in a MO setting)--for instance, in the context of low-impact AI \cite{vamplew_potential-based_2021}, balancing across a safety objective
and a primary objective. A leximin approach orders a set of objectives, and then optimizes for the first value in the set, followed by the second value, and so on; a formal description can be found in \cite{vamplew_human-aligned_2018}.

\subsubsection{Non-linear multiple objective functions}
Non-linear utility functions have been previously explored in \cite{rolf_need_2020}. It was found that a non-linear objective system traversing a learning-space through reinforcement learning learns highly satisfactory solutions, balancing contradictory needs. That work followed earlier approaches that attempted to exhaustively explore \cite{van2014multi,parisi2016multi} a space or a subset thereof \cite{barrett2008learning} of Pareto-improvements to the current state space.

A multiple objective reward exponential function was proposed \cite{rolf_need_2020}, of the form:

\begin{align}\label{eq:rolf}
f(x)= &  -\exp(-x) \\ \nonumber
\end{align}

 where $x$ is untransformed utility signal, and $f(x)$ is a function that creates a `loss averse' transformation of the utility.

The methods section below introduces alternative multiple objective exponential functions and explains the bases for the deviations from the previously proposed \cite{rolf_need_2020} design as in Equation~\ref{eq:rolf}.

\subsubsection{AI Morality}

There has been at least one prior effort made to capture moral uncertainty in AI \cite{martinho_empirical_2020}. In this project, a discrete choice analysis model was used to demonstrate moral uncertainty about alternative policy choices.



\subsubsection{Theoretical approaches}

`Conservative agency' has been previously described as a unification of side effect avoidance, state change minimization, and reachability preservation \cite{armstrong_low_2017, turner_conservative_2020}. Its goal is to optimize `the primary reward function while preserving the ability to optimize others', or `Attainable Utility Preservation'.

Conservativism in Bayesian \cite{pmlr-v125-cohen20a} or neuromorphic systems \cite{byrnes_steve_conservatism_2020} has also been previously proposed, including the possibility of requesting help from an agent mentor.

\subsection{Building on previous work}

This paper is the first to examine continuous non-linear multi-objective decision-making in the context of low-impact AI work as described in \cite{vamplew_potential-based_2021}. It is also the first we are aware of to apply a split-function exponential-log transform to any AI decision-making or RL application.  Previous work has explored thresholded functions for traversing environments where rewards need to be gathered in different parts of the environment and traded off over time \cite{rolf_need_2020}, although that work focused on a function resembling ELA and did not explore SFELLA. 

\cite{turner_conservative_2020} started out with similar goals to ours; they described `conservative agency' to balance `optimization of the primary reward function with preservation of the ability to optimize auxiliary reward functions'. They did not examine non-linear combinations of objectives, and instead focused on learning approaches for optimizing the scaling between objectives. We have not applied arbitrary scaling between objectives, and applying the scaling method as in \cite{turner_conservative_2020} could be complementary to our work.

\subsection{Pluralistic human value system}

Often, AI alignment aims to ensure AI systems fulfill human preferences. While neither human preferences nor human values are always consistent \cite{sotala2016defining}, values are higher-order and harder to identify \cite{barrett2008learning}, but preferences are more sensitive to context and recalculation \cite{warren2011values}. The framework here focuses on modeling distinct human values as distinct objectives, while recognizing that there may be many preferences to satisfy within each overarching value function. As outlined above, intuitions of individuals frequently conflict \cite{haidt2001emotional} and moral views between individuals also conflict \cite{bogosian_implementation_2017}.

It has been argued that one way to address uncertainty in moral decision-making is to learn human moral judgement in a bottom-up fashion \cite{bogosian_implementation_2017}; rather than learning human values, an agent learns human preferences, and those preferences are implicitly held within values. Even if this is technically adequate, in practice it might be necessary to put constraints on system to ensure they don't learn anti-social preferences \cite{neff2016talking}.


Furthermore, a utility function based on human preferences themselves has been argued to be an insufficient definition of value \cite{sotala2016defining, DBLP:journals/corr/abs-1712-05812}, because \begin{enumerate*}
    \item humans do not have consistent utility functions,
    \item utility functions are poor models of conflicts between lower- and higher-order preferences,
    \item it fails to draw distinctions between `wanting' and `liking', and
    \item a utility function of unitary value could not adequately generalize from existing values to new ones.
\end{enumerate*}
It is arguably also an important question of how to combine the rewards that are based on human preferences. The proper way might not be a trivial sum of the individual rewards since that would skip the nonlinear transformation by the utility functions before the final aggregation takes place. 

A theoretical Bayesian preference-learning system could model preferences and through learning human values, learn the proper way to combine them. But there is the trade-off between a model being too simple (linear sum of rewards), and too complex (a Bayesian network, requires potentially unpractical amounts of data). The middle ground would be to have a model-based approach which describes some rules (like the presence of negative exponential shape for violated alignment objectives) while being still flexible and able to learn the data as parameters of the model.



\subsection{Design principles}
The following principles guided us in selecting an aggregate function different to the maximin or leximin approaches:
\begin{itemize}
    \item \textit{Loss aversion}, \textit{conservatism}, or \textit{soft maximin}. We seek to improve the position of the lowest member of the set of values, while also not entirely disregarding optimization of other values.
    \item \textit{Balancing outcomes across objectives}. 
    We are concerned about moral-system and human-values applications of multi-objective systems, where each objective represents a different moral system or value. Each moral system or value bears some value, but no precise equivalence or conversion rate between them can be determined. To be conservative and ensure a low probability of any bad outcome, we avoid strongly negative outcomes in terms of any objective. Alternatively, each objective represents a particular subject's preferences. Then, balancing outcomes across objectives represents an implementation of fairness between subjects.
    \item \textit{Zero-point consistency}. An agent evaluates whether an action performs better not only compared to alternatives, but also compared to no action at all, which would have a value of 0. For this reason any aggregation or transformation function should preserve the overall estimated sign or valence of an objective.
\end{itemize}

Previous work \cite{vamplew_potential-based_2021} has described thresholded leximin approaches in order to trade-off objectives, in the context of trading off a primary objective and an impact Objective in low-impact AI. A thresholded leximin function aims to first maximize the thresholded value of thresholded objectives, and then secondarily maximize the unthresholded value of one or more other objectives. If the alignment objective is thresholded, then the system aims to first achieve at least a thresholded level of the alignment objective, and then subject to this, to achieve a maximum level of the performance objective. Alternatively, a \textit{complete thresholded leximin}, aims to maximize the thresholded value of all objectives, i.e., reach the threshold on each objective; then, subject to this, aims to maximize the unthresholded value of each objective.

This complete thresholded leximin is a discretely-stepped maximin approximation. Reaching a specified minimum threshold value on each objective takes precedence over maximizing already-high values. Yet it is not a strict maximin, because the function doesn't only care about maximizing the minimum value; in fact, beyond a specified threshold, no value is given at all. In this way a thresholded leximin can be seen as a compromise between a maximin function and a linear maximum expected utility (MEU) function.

In this paper we propose another compromise between a maximin and a linear MEU function: here, following previous work \cite{rolf_need_2020}, we propose a continuous rather than discrete trade-off between maximin and linear MEU. This approach avoids specifying a threshold, which may be desirable for at least three reasons. First, it might not be possible to specify an appropriate threshold in advance. Second, continuously decreasing the extent to which we prioritize an objective might better fit our underlying aims or values than giving a high priority up to a threshold and no priority at all above that threshold. 
Third, in the context of modeling human values, this approach might sometimes be more consistent with human value processing\cite{Tom515}, considering the literature on risk aversion \cite{pratt1978risk}.

A continuous compromise between multiple objectives also offers greater benefits for complex low-impact artificial systems. If one had dozens of objectives, a strict maximin or leximin function might come to be overly inflexible. 

\section{Experiment 1}
\subsection{Method}
We adapted algorithms in \cite{vamplew_potential-based_2021}, comparing the existing \tloA{} aggregation function with several new aggregation and scalarization functions. These were compared using the same environments and benchmarks as in \cite{vamplew_potential-based_2021}. Specifically, the environments were the UnbreakableBottles (UB) environment, the BreakableBottles (BB) environment, the Sokoban environment, and the Doors environment.

For replication purposes, a complete set of the code used to generate the experiments presented here, as well as this paper itself, can be found at \url{https://github.com/bjsmith/multi-objective-value-aggregation/tree/e1a3acc94b59d7cda25842c5b3dfdefcf1f91685}.

\subsection{Testing environment}

We used a modified version of the testing environment from \cite{vamplew_potential-based_2021}, obtained with permission from the authors. Learning proceeded for 5,000 episodes, and performance during learning (online performance) and after learning (offline performance) was measured. There were several key changes for our implementation. Changes affecting modeling results themselves are described below. Each experimental condition itself run for 100 trials in order to build up a probability distribution, making the comparison results generalisable to repeated runs of the task.

\subsubsection{Aggregation functions}

\begin{figure*}[h]
 
  \includegraphics[width=\columnwidth]{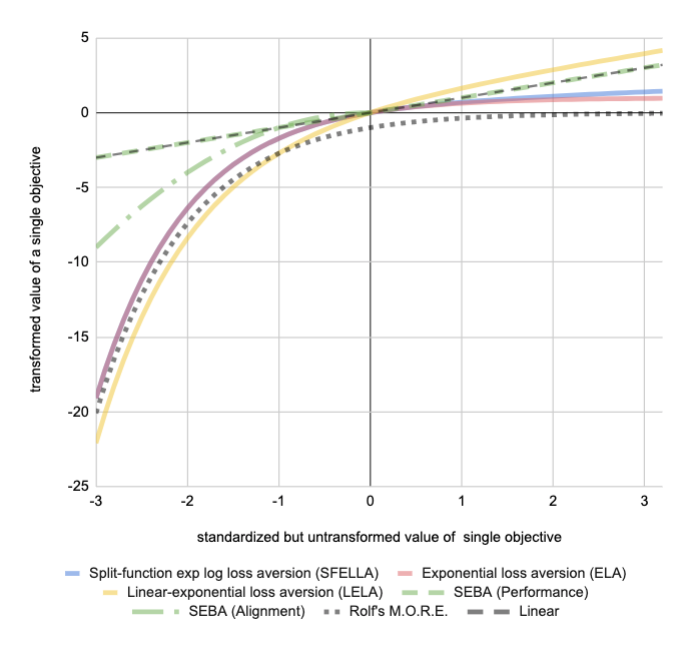}
  \includegraphics[width=\columnwidth]{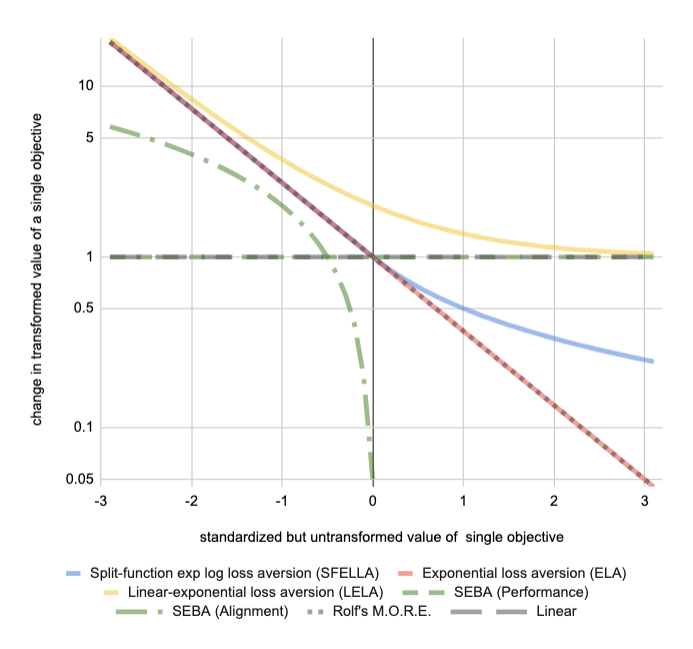}
  \caption{Transform functions. Left: Each transform function is applied to the reward received from the environment for each objective, or to the Q value of the RL agent for each objective. In our current setup it is applied to the Q values of the RL agent.
  The output of each of these transform functions are averaged together 
  (Equation~\ref{eq:meu}). 
  Right: Change in $f(x)$ per unit $x$, with $y$ axis plotted on a log scale
  . Note that ELA and SFELLA produce greater-than-linear change in $f(x)$ when $x<0$ and less-than-linear change when $x>0$. In contrast, LELA's change never falls below 1.}
  \label{fig:transform_functions}
  \Description{Transform functions.}
\end{figure*}

\begin{figure*}[h]
 
  \includegraphics[width=\columnwidth]{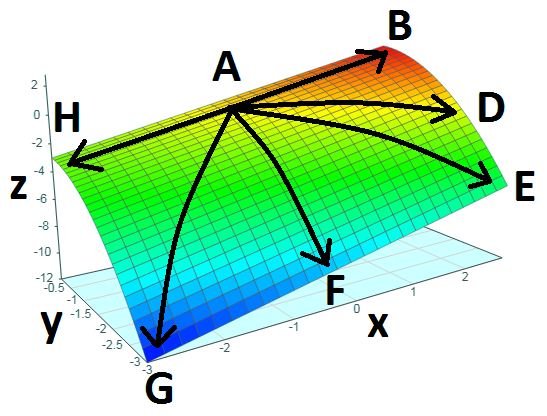}
  \includegraphics[width=\columnwidth]{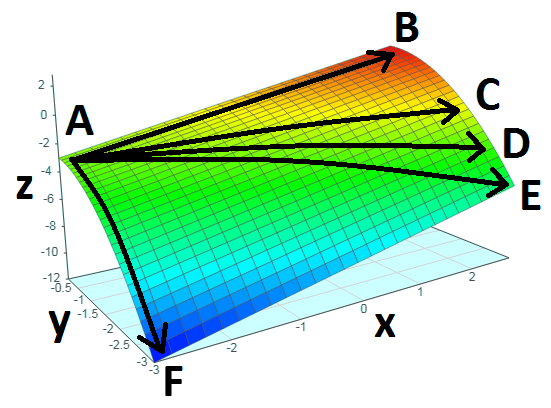}
  \caption{Transform for the SEBA aggregation. One of the objectives is the alignment objective (the y-axis), and the other objective is the performance objective (the x-axis). The z-axis represents the aggregated utility. The two types of objectives are treated differently. The performance objective has always linear treatment regardless of the current sign of its input measure, while the alignment measure is upper-bounded at zero and has exponential treatment (in case of SEBA it is a negated squared error). These plots illustrate two things. 1. The performance objective (the x-axis) is linear regardless of the sign of the value of the input measure. 2. The alignment related measure (the y-axis) may be sacrificed, but only up to a degree. Once alignment would be sacrificed too much, the evaluation of the aggregated utility quickly becomes strongly negative, since the alignment measure is treated exponentially. So this provides the loss aversion aspect.}
  \label{fig:seba_transform_functions_3d}
  \Description{Transform for the SEBA aggregation.}
\end{figure*}

All of the multi-objective utility functions we compared use the following steps: 
\begin{enumerate}
    \item 
    A scaling factor $c_i$ is applied to each objective, specific to that reinforcement, by multiplying it with the value of reinforcement at each time point. This step has not been implemented in this paper but we emphasize its possible use in the future. 
    \item Transform the scaled output by using a non-linear transform, as described below (Figure~\ref{fig:transform_functions} and Figure ~\ref{fig:seba_transform_functions_3d}). 
    
    \item Combine the transformed output using a simple average/sum as in Equation~\ref{eq:meu}.
\end{enumerate}

These steps describe a utility function which can be applied either to the individual reinforcement delivered at each in response to an action, or to Q-values during action selection. In Experiment 1, we apply these steps to Q-values during action selection.

Each non-linear function is applied component-wise before aggregation occurs by averaging/summing of the transformed values

\begin{align}
\label{eq:meu}
U=\sum_{i}^n{f_i(c_i x_i)}
\end{align}

\noindent where ${f_i}$ describes a specific transform function for the $i$th objective, explained in more detail below. Note that here, $U$ describes our modeling of Q-values. The scaling factors $c_i$ can be treated as parameters of the model. They could be, for example, specified directly by the human, automatically determined in the future by the agent using a value learning method, or calculated by some other algorithm. 
For the purposes of this experiment, we left these at $c=1$ (instead, we directly modified $x$, the value returned by the environment), but emphasize that modifying these scale values could be useful in the future.

 New non-linear transforms compared are:

\begin{itemize}
    \item Split-function exp-log loss aversion (SFELLA)
    \item Exponential loss aversion (ELA)
    \item Linear-exponential loss aversion (LELA)
    \item Squared error based alignment (SEBA)
\end{itemize}

\noindent The SEBA transform function envisages differing functions for `primary' and `alignment' categories of objectives. All other transform functions do not distinguish categories of objectives, and apply the same function over all objectives.

Each non-linear transform is a transform of the value obtained along a specific objective 
at a specific state 
with a specific action. 
The SFELLA, ELA, and LELA functions are illustrated in Figure~\ref{fig:transform_functions}. The SEBA aggregation is illustrated on Figure ~\ref{fig:seba_transform_functions_3d}.

For each transform, where $x=0$, $f(x)=0$. This is a minor modification from the non-linear transform previously proposed in \cite{rolf_need_2020}, typically achieved by adding 1 to the outcome value.

Each function also provides that $\frac{\mathrm{d} f(x) }{\mathrm{d} x}$ declines as $x$ gets larger. This lowers inequality between outcomes as measured in different objectives, objectives where values are strongly negative get disproportionately higher priority. Where different objectives were operationalizing, for instance, priorities among different interested parties, this might be particularly useful in reducing inequality between outcomes.

In SFELLA, there is a split in the function at $x=0$. It expresses a loss-averse function where losses will be amplified more than gains:

\begin{align}
f(x)= & \ln(cx+1) & \mathrm{ where \: x>0} \\ \nonumber
  &  -\exp(-cx)+1 &  \mathrm{otherwise} \\ \nonumber
\end{align}

Additionally, by implementing a log rather than a negative exponential in the positive domain, the function retains relatively more weight on positive objectives, i.e. is not bounded.

The ELA is a simplification of this without a case distinction at the cost of giving very little weight to any increase in values over 1:

\begin{align}
f(x)= &  -\exp(-cx)+1 \\ \nonumber
\end{align}

With LELA we add a $x$ term so that value continues to increase at least linearly for large inputs:

\begin{align}
f(x)= &  -\exp(-cx)+cx+1 \\ \nonumber
\end{align}

This still yields loss aversion at points less than zero but always provides that an increase in $x$ increases at least linearly in $f(x)$

Finally, SEBA takes a different approach in that rather than treating each objective identically, transformations are applied differently to performance and alignment objectives.

For performance objectives the SEBA formula is linear:
\begin{align}
f(x)= &  cx \\ \nonumber
\end{align}
There is no differentiation between negative and positive areas of the measures of the performance objectives. This avoids the need for establishing a zero-point. Proper scaling is still needed.

SEBA expresses loss aversion for alignment objectives using a negated square power function, and assumes that alignment objectives are non-positive:
\begin{align}
\label{eq:seba}
f(x)= &  -(cx)^2 \\ \nonumber
  &  \mathrm{ where \: x \leq 0}
\end{align}

The alignment related measures still have a “natural” zero-point, since they by definition are bounded at zero where no (soft) constraint violations are occurring. 
Such measures would usually measure the deviation of something from a desired target value. Such measures have two main types:
\begin{itemize}
    \item The desired target value is zero (for example, zero harm, etc).
    \item Alternatively it might be a homeostatic set-point (for example, optimal temperature, etc), so the measure is representing the negated absolute value of the deviation regardless of the direction of the deviation.
\end{itemize}
The SEBA aggregation is illustrated on Figure ~\ref{fig:seba_transform_functions_3d}. A number of specific situations are illustrated in the graph using upper-case letter points, and it is helpful to consider their interpretation:
  \begin{itemize}
      \item A - The initial state. The alignment objective / soft constraint is met and the performance objective is either at zero (left side plot) or at negative value (right side plot).
      \item B - The performance objective is improved, the alignment constraint is preserved. Moving in this direction changes the aggregated score linearly thus enabling independence from the zero-point.
      \item C - Shown only on the right side plot. Performance objective is improved significantly, while alignment constraint is sacrificed just so slightly that the aggregated utility is still improved.
      \item D - Performance objective is improved significantly, but the alignment constraint is sacrificed so much that the aggregated utility does not change as compared to the initial state. The agent is neutral to this state change and is not driven towards this state nor avoiding it.
      \item E - Performance objective is improved significantly, while the alignment constraint is violated significantly. Therefore the aggregated utility becomes worse than the initial state. The agent avoids this state.
      \item F - The measure for the performance objective does not change, but the alignment constraint gets violated.
      \item G - Shown only on the left side plot. Both the performance objective and the alignment objective / constraint get worse.
      \item H - Shown only on the left side plot. The performance objective gets much worse but the alignment constraint is still satisfied. It is also noteworthy that this state is evaluated to be about as good as the alternative state somewhere between D and E where the alignment constraint is getting notably violated but the performance objective is improved much.
  \end{itemize}

\subsubsection{Environment}

\begin{figure}
    \centering
    \includegraphics[width=1\columnwidth]{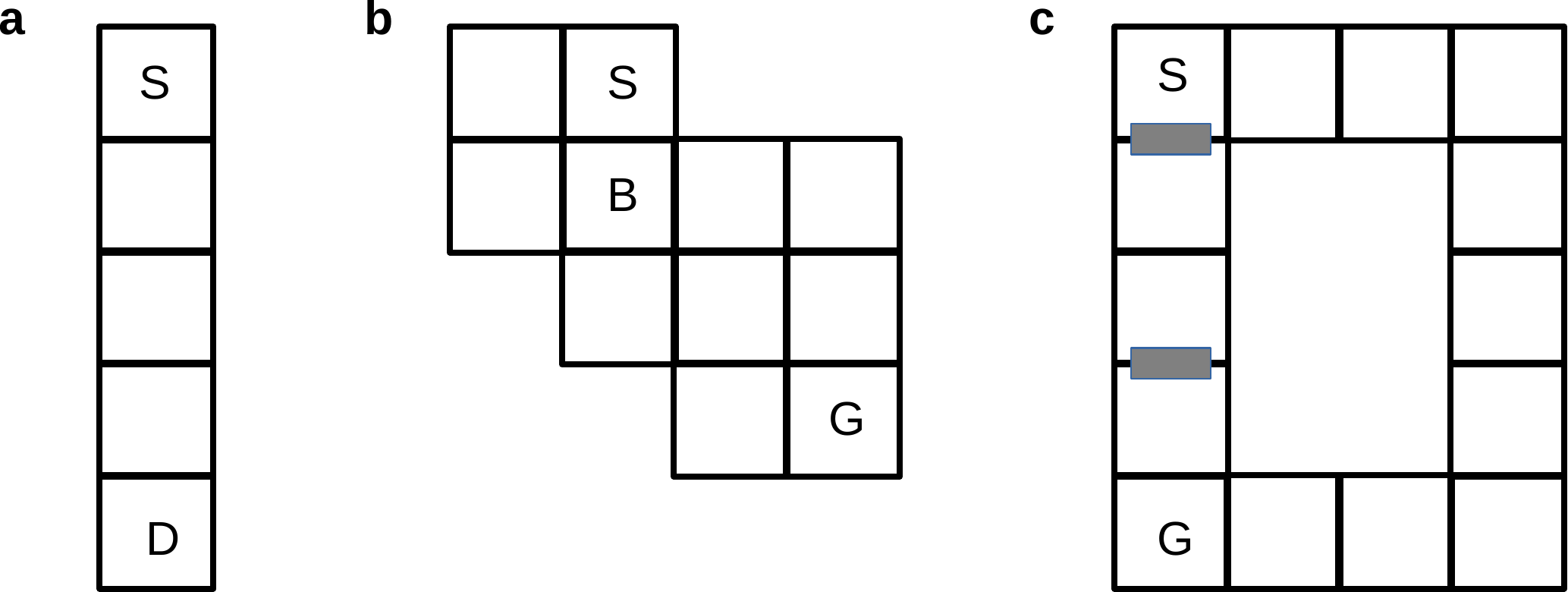}
    \caption{Maps of physical environments in (a) (Un)Breakable Bottles, (b) Sokoban, (c) Doors. Based on Figure in \cite{vamplew_potential-based_2021}.}
    \label{fig:envs}
\end{figure}

In every environment, agents have two objectives: a `performance' objective (denoted \RP{}) and an `alignment' objective (denoted \RA{}). 
Four gridworld environments reported in \cite{vamplew_potential-based_2021} were examined.
They are shown in Figure \ref{fig:envs} and we call them the 'Breakable and Unbreakable Bottles' (BB and UB), 'Sokoban' and 'Doors'.

The Bottles environments share the same 1D grid layout where one end is the destination 'D' where the agent has to deliver bottles and the other end is the source 'S' where bottles are provided.
Initially, the agent does not carry a bottle and it can hold up to two bottles and an episode ends when two bottles have been delivered.
While in between source and destination an agent holding two bottles can drop a bottle on a tile with a probability of 10\%.
Leaving a bottle on the way yields a penalty of -50 in R$^*$.
While in UnbreakableBottles the bottles can be picked up again where they were left, in BreakableBottles they break upon dropping hence irreversibly changing the environment and receiving the penalty.

In the Sokoban environment the agent starts on tile 'S' and is tasked with pushing away the box 'B' in order to reach the goal tile 'G'.
There are two ways of pushing: downwards into a corner (irreversible) and to the left (reversible, but involving more steps).
A penalty of -25 is evoked for each wall touching the box in the final position.

In the Doors environment the agent must simply travel from the start 'S' to the goal 'G'.
It can choose to open or close the doors (grey) which takes each one action. 
There are two possible paths: either the agent can move around the right corridor taking 10 moves to reach 'G' or the agent can move straight down by opening the doors (6 moves if the doors stay open).
However, there is a penalty of -10 associated with leaving a door open.
Therefore the desired solution is moving down while closing the doors behind the agent taking 8 moves.

We wanted to understand how different aggregation functions could respond to re-scaling of primary or alignment rewards. To do this, we repeated each experiment 9 times. The first time was with the original settings as in \cite{vamplew_potential-based_2021}. Then, we repeated this with each environment's primary objective feedback scaled by $10^{-2}$, $10^{-1}$, $10^1$, and $10^2$. The same range of scaling was then applied to the alignment objective feedback. This scaling could in some scenarios potentially be distinguished from $c$ in Equations~\ref{eq:meu}-\ref{eq:seba}. Even though it is mathematically equivalent in our implementation, it could represent changes to the environment rather than changes in agent evaluation.

\subsubsection{Measurement}

Each of the proposed functions was compared against the best performing function in \cite{vamplew_potential-based_2021}, the \tloA{} function, on the `R$^*$' metric from the same paper. The `R$^*$' arbitrarily scores a weighted combination of primary and alignment objectives where one unit of alignment objective (always on a negative scale) is worth 10-50 units of the primary objective, depending on the environment.

Learning was switched off after 5000 episodes. Following that time, offline performance was observed as the average across offline performance in all 100 runs.

\subsection{Results}
In offline performance, there was very little difference between the best-performing proposed function and \tloA{}. For this reason, the remainder of the results reported will discuss performance during online testing, i.e., performance during learning itself.

\begin{table}[t]
\footnotesize
  \caption{Mean $\text{R}^*$ Online performance. Each row represents comparable performance across 5 different objective functions. Values within 10\% of the best value in each row are highlighted. Higher scores are better. Items are significantly different from \tloA{} when marked *$p<0.05$, ** $p <0.01$, *** $p<0.001$; arrows mark the direction of significant differences.}
  \label{tab:mean_r_star_performance}
\begin{adjustbox}{width=\columnwidth}

\begin{tabular}{>{\raggedright\arraybackslash}p{5em}>{\raggedleft\arraybackslash}p{4em}>{\raggedright\arraybackslash}p{4.5em}rrrr}
\toprule
Environment & Objective Modified & Objective Scale & SEBA & SFELLA & LinearSum & TLO$^A$\\
\midrule
 &  & 1 & \textcolor{black}{1.43$\downarrow$**} & \textcolor{blue}{6.54$\uparrow$***} & \textcolor{black}{1.48$\downarrow$*} & \textcolor{black}{1.81}\\
\cmidrule{2-7}
 &  & 0.01 & \textcolor{black}{1.33} & \textcolor{black}{1.38} & \textcolor{black}{1.47} & \textcolor{black}{1.46}\\

 &  & 0.1 & \textcolor{black}{1.39} & \textcolor{black}{1.88$\uparrow$**} & \textcolor{black}{1.37} & \textcolor{black}{1.41}\\

 &  & 10 & \textcolor{blue}{6.32$\uparrow$***} & \textcolor{black}{4.44$\uparrow$***} & \textcolor{black}{5.61$\uparrow$***} & \textcolor{black}{-0.22}\\

 & \multirow[t]{-4}{4em}{\raggedleft\arraybackslash Alignment} & 100 & \textcolor{black}{2.22$\uparrow$***} & \textcolor{black}{-3.49$\downarrow$***} & \textcolor{blue}{6.05$\uparrow$***} & \textcolor{black}{-0.48}\\
\cmidrule{2-7}
 &  & 0.01 & \textcolor{blue}{6.34$\uparrow$***} & \textcolor{black}{5.51$\uparrow$***} & \textcolor{blue}{6.01$\uparrow$***} & \textcolor{black}{1.96}\\

 &  & 0.1 & \textcolor{black}{2.46$\uparrow$**} & \textcolor{blue}{6.43$\uparrow$***} & \textcolor{black}{5.43$\uparrow$***} & \textcolor{black}{1.88}\\

 &  & 10 & \textcolor{black}{1.41$\downarrow$**} & \textcolor{blue}{6.51$\uparrow$***} & \textcolor{black}{1.44$\downarrow$*} & \textcolor{black}{1.77}\\

\multirow[t]{-9}{5em}{\raggedright\arraybackslash BB} & \multirow[t]{-4}{4em}{\raggedleft\arraybackslash Primary} & 100 & \textcolor{black}{1.46$\downarrow$**} & \textcolor{blue}{6.40$\uparrow$***} & \textcolor{black}{1.35$\downarrow$***} & \textcolor{black}{1.81}\\
\cmidrule{1-7}
 &  & 1 & \textcolor{black}{-0.48$\downarrow$***} & \textcolor{black}{4.38$\uparrow$***} & \textcolor{black}{-0.47$\downarrow$***} & \textcolor{black}{3.87}\\
\cmidrule{2-7}
 &  & 0.01 & \textcolor{black}{-0.73$\downarrow$*} & \textcolor{black}{-0.58} & \textcolor{black}{-0.48} & \textcolor{black}{-0.49}\\

 &  & 0.1 & \textcolor{black}{-0.64} & \textcolor{blue}{8.29$\uparrow$***} & \textcolor{black}{-0.52} & \textcolor{black}{-0.63}\\

 &  & 10 & \textcolor{black}{3.43} & \textcolor{black}{3.74} & \textcolor{blue}{5.75$\uparrow$***} & \textcolor{black}{3.63}\\

 & \multirow[t]{-4}{4em}{\raggedleft\arraybackslash Alignment} & 100 & \textcolor{black}{3.16$\uparrow$**} & \textcolor{black}{2.73} & \textcolor{black}{3.95$\uparrow$***} & \textcolor{black}{2.82}\\
\cmidrule{2-7}
 &  & 0.01 & \textcolor{black}{3.43$\downarrow$***} & \textcolor{black}{3.66$\downarrow$***} & \textcolor{black}{4.05} & \textcolor{black}{4.09}\\

 &  & 0.1 & \textcolor{black}{5.39$\uparrow$***} & \textcolor{black}{4.10} & \textcolor{blue}{5.71$\uparrow$***} & \textcolor{black}{3.91}\\

 &  & 10 & \textcolor{black}{-0.70$\downarrow$***} & \textcolor{blue}{4.41$\uparrow$***} & \textcolor{black}{-0.67$\downarrow$***} & \textcolor{blue}{3.97}\\

\multirow[t]{-9}{5em}{\raggedright\arraybackslash Doors} & \multirow[t]{-4}{4em}{\raggedleft\arraybackslash Primary} & 100 & \textcolor{black}{-0.51$\downarrow$***} & \textcolor{blue}{4.17$\uparrow$**} & \textcolor{black}{-0.58$\downarrow$***} & \textcolor{blue}{3.85}\\
\cmidrule{1-7}
 &  & 1 & \textcolor{black}{-14.98$\downarrow$***} & \textcolor{black}{-10.29$\downarrow$***} & \textcolor{black}{-14.97$\downarrow$***} & \textcolor{blue}{10.76}\\
\cmidrule{2-7}
 &  & 0.01 & \textcolor{black}{-15.02} & \textcolor{black}{-14.97} & \textcolor{black}{-14.98} & \textcolor{black}{-14.97}\\

 &  & 0.1 & \textcolor{black}{-14.96} & \textcolor{black}{-14.98} & \textcolor{black}{-14.99} & \textcolor{black}{-14.95}\\

 &  & 10 & \textcolor{blue}{10.88$\uparrow$***} & \textcolor{blue}{10.92$\uparrow$***} & \textcolor{black}{-14.95$\downarrow$***} & \textcolor{blue}{10.72}\\

 & \multirow[t]{-4}{4em}{\raggedleft\arraybackslash Alignment} & 100 & \textcolor{blue}{10.82$\uparrow$***} & \textcolor{black}{3.76$\downarrow$***} & \textcolor{blue}{10.86$\uparrow$***} & \textcolor{blue}{10.49}\\
\cmidrule{2-7}
 &  & 0.01 & \textcolor{blue}{10.91$\uparrow$***} & \textcolor{blue}{10.86$\uparrow$*} & \textcolor{blue}{10.82} & \textcolor{blue}{10.77}\\

 &  & 0.1 & \textcolor{black}{-14.96$\downarrow$***} & \textcolor{black}{5.97$\downarrow$***} & \textcolor{black}{-14.95$\downarrow$***} & \textcolor{blue}{10.82}\\

 &  & 10 & \textcolor{black}{-15.01$\downarrow$***} & \textcolor{black}{-11.05$\downarrow$***} & \textcolor{black}{-14.98$\downarrow$***} & \textcolor{blue}{10.88}\\

\multirow[t]{-9}{5em}{\raggedright\arraybackslash Sokoban} & \multirow[t]{-4}{4em}{\raggedleft\arraybackslash Primary} & 100 & \textcolor{black}{-14.96$\downarrow$***} & \textcolor{black}{-10.97$\downarrow$***} & \textcolor{black}{-14.97$\downarrow$***} & \textcolor{blue}{10.82}\\
\cmidrule{1-7}
 &  & 1 & \textcolor{blue}{28.71$\uparrow$***} & \textcolor{blue}{27.99$\uparrow$***} & \textcolor{blue}{28.76$\uparrow$***} & \textcolor{blue}{27.09}\\
\cmidrule{2-7}
 &  & 0.01 & \textcolor{blue}{28.70} & \textcolor{blue}{28.73} & \textcolor{blue}{28.74} & \textcolor{blue}{28.79}\\

 &  & 0.1 & \textcolor{blue}{28.72} & \textcolor{blue}{28.74} & \textcolor{blue}{28.77} & \textcolor{blue}{28.72}\\

 &  & 10 & \textcolor{blue}{27.62$\uparrow$***} & \textcolor{blue}{25.90$\uparrow$***} & \textcolor{blue}{28.72$\uparrow$***} & \textcolor{black}{23.37}\\

 & \multirow[t]{-4}{4em}{\raggedleft\arraybackslash Alignment} & 100 & \textcolor{blue}{25.66$\uparrow$***} & \textcolor{black}{18.67$\uparrow$***} & \textcolor{blue}{27.42$\uparrow$***} & \textcolor{black}{14.60}\\
\cmidrule{2-7}
 &  & 0.01 & \textcolor{blue}{27.73$\uparrow$***} & \textcolor{blue}{26.79$\downarrow$*} & \textcolor{blue}{27.31$\uparrow$***} & \textcolor{blue}{26.98}\\

 &  & 0.1 & \textcolor{blue}{28.66$\uparrow$***} & \textcolor{blue}{27.82$\uparrow$***} & \textcolor{blue}{28.64$\uparrow$***} & \textcolor{blue}{27.15}\\

 &  & 10 & \textcolor{blue}{28.78$\uparrow$***} & \textcolor{blue}{27.91$\uparrow$***} & \textcolor{blue}{28.69$\uparrow$***} & \textcolor{blue}{27.10}\\

\multirow[t]{-9}{5em}{\raggedright\arraybackslash UB} & \multirow[t]{-4}{4em}{\raggedleft\arraybackslash Primary} & 100 & \textcolor{blue}{28.75$\uparrow$***} & \textcolor{blue}{27.85$\uparrow$***} & \textcolor{blue}{28.71$\uparrow$***} & \textcolor{blue}{27.08}\\
\bottomrule
\end{tabular}

\end{adjustbox}
\end{table}

 


\begin{figure}
  \includegraphics[width=\columnwidth]{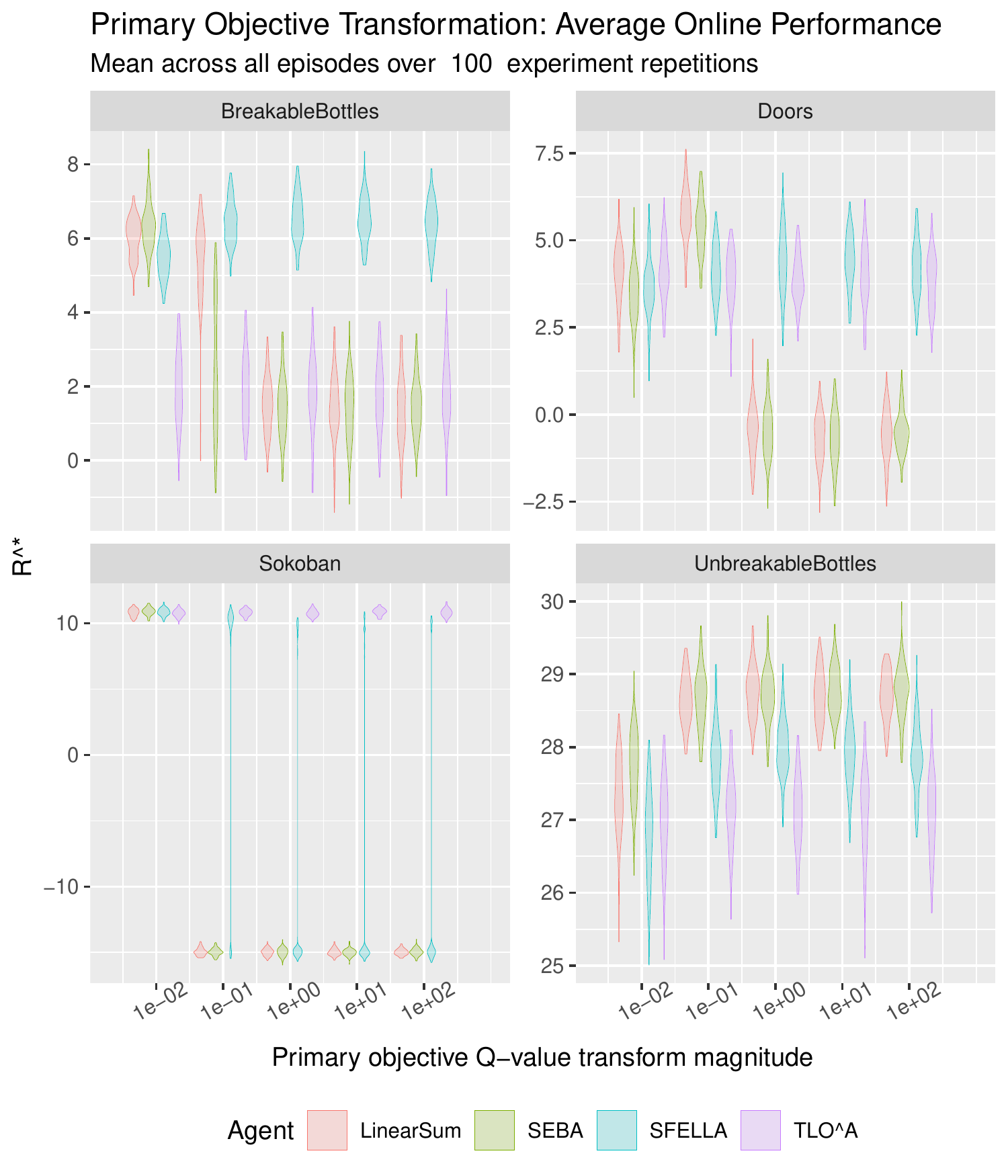}
  \includegraphics[width=\columnwidth]{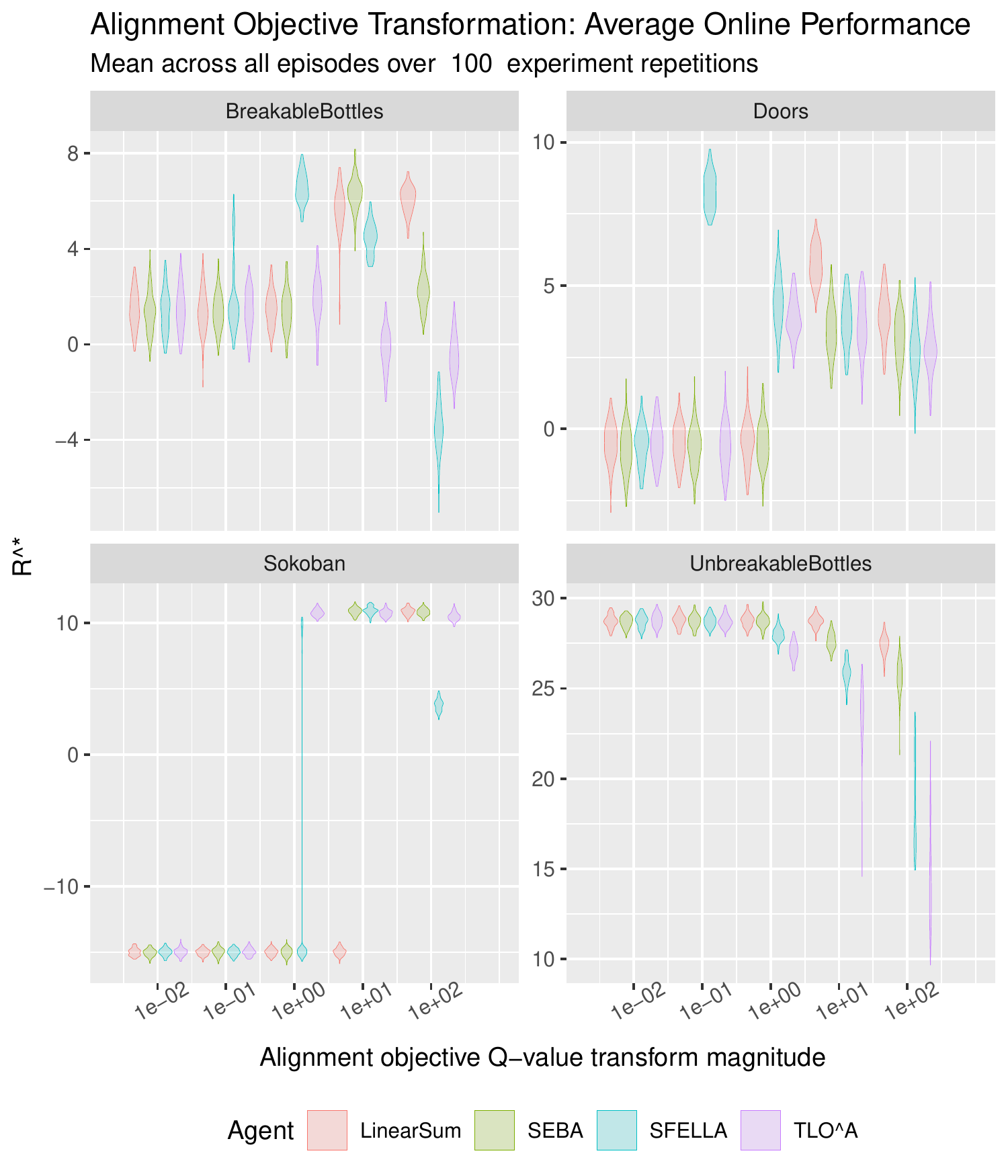}
  \caption{Experiment 1: \RStar{} Online performance averaged across learning episodes and experiment repetitions for different Q-value transforms. (A): \RStar{} when scaling primary Q-values across 5000 learning trials. SFELLA consistently performs similar or better to \tloA{}. (B): \RStar{} when transforming alignment Q-values across 5000 learning trials. No algorithm is a clear best performer.}
   \label{fig:online_performance}
 \end{figure}

While there was no clear best performer, SFELLA had the best online performance during training across a wider range of environments and environment variants than any other agent, including \tloA{}. Table~\ref{tab:mean_r_star_performance} describes relative $\text{R}^*$ scores for each function, compared to the $\text{TLO}^\text{A}$ function, at different scales.  Within the BreakableBottles environment, \tloA{} performed worse than all other agents at all scales, and SFELLA performed within 10\% of the best within five of nine scales. In the UnbreakableBottles Environment, performance between all agents except ELA was not significantly different. Agents in the Sokoban environment were generally very sensitive to scaling, though \tloA{} performed better in this environment overall.

SFELLA tended to perform at best level when re-scaling the primary objective (Figure~\ref{fig:online_performance}a), but less well when alignment objective was re-scaled (Figure~\ref{fig:online_performance}b).

\section{Experiment 2}

Transforming rewards $r_t$ rather than transforming Q-values $Q(s, a)$ might represent a different challenge for the models we presented. Given repeated occurrences of the same $s, a$ learning pair, there's no long-run difference between the two because they both approach the same `learning asymptote'. However, during the learning process, each transformation process produces different rates of learning for positive and negative domains, under conditions described in more detail below. Faster rates of learning are not always advantageous. At least two observations should be made.



First, speed of learning is affected differently within positive and negative domains. In the positive domain, transformation on reward lowers the speed of learning, while transformation on Q-value lowers the asymptote of learning without directly reducing speed of learning. Thus, in the positive domain, transformation on utility leads to slower learning. Conversely, in the negative domain, transformation on reward exaggerates the speed of learning, so transformation on utility leads to faster learning in the negative domain, while the transformation on Q-value exaggerates only the (negative) asymptote of learning without directly affecting learning speed.  Note that they will all approach the same asymptote in the end.

Second, this has significant implications for differences in the granularity of rewards. Consider two utility schedules. In the first schedule, every 1 timestep an agent is given a small penalty--such as the time penalty given in the BreakableBottles task for every timestep the challenge hasn't been completed. In the second schedule, an agent occasionally receives a large penalty for an action a at $s, a$, such as pushing a box into a corner. Because learning is sparser, the slower learning inherent in reward transformation (in the positive domain) and Q-value transformation (in the negative domain) could actually substantially influence behavior for a substantial part of the 5000 episodes over online learning.

Very generally, speeding up learning in the negative domain is relatively more cautious, while speeding up learning in the positive domain is relatively less cautious. For small magnitudes, transformation makes little difference. Of the two transformation processes, for large magnitudes, reward tranformation produces a more cautious outcome than Q-value, and responds much more strongly to occasional strongly negative feedback than to regular slightly negative feedback.




\subsection{Method}

We repeated the experiment, transforming rewards rather than transforming Q-values. All of the same parameters applied as in Experiment 1, but \RP{} and \RA{}, rather than Q-values were transformed.

\subsection{Results}

\begin{figure}
  \includegraphics[width=\columnwidth]{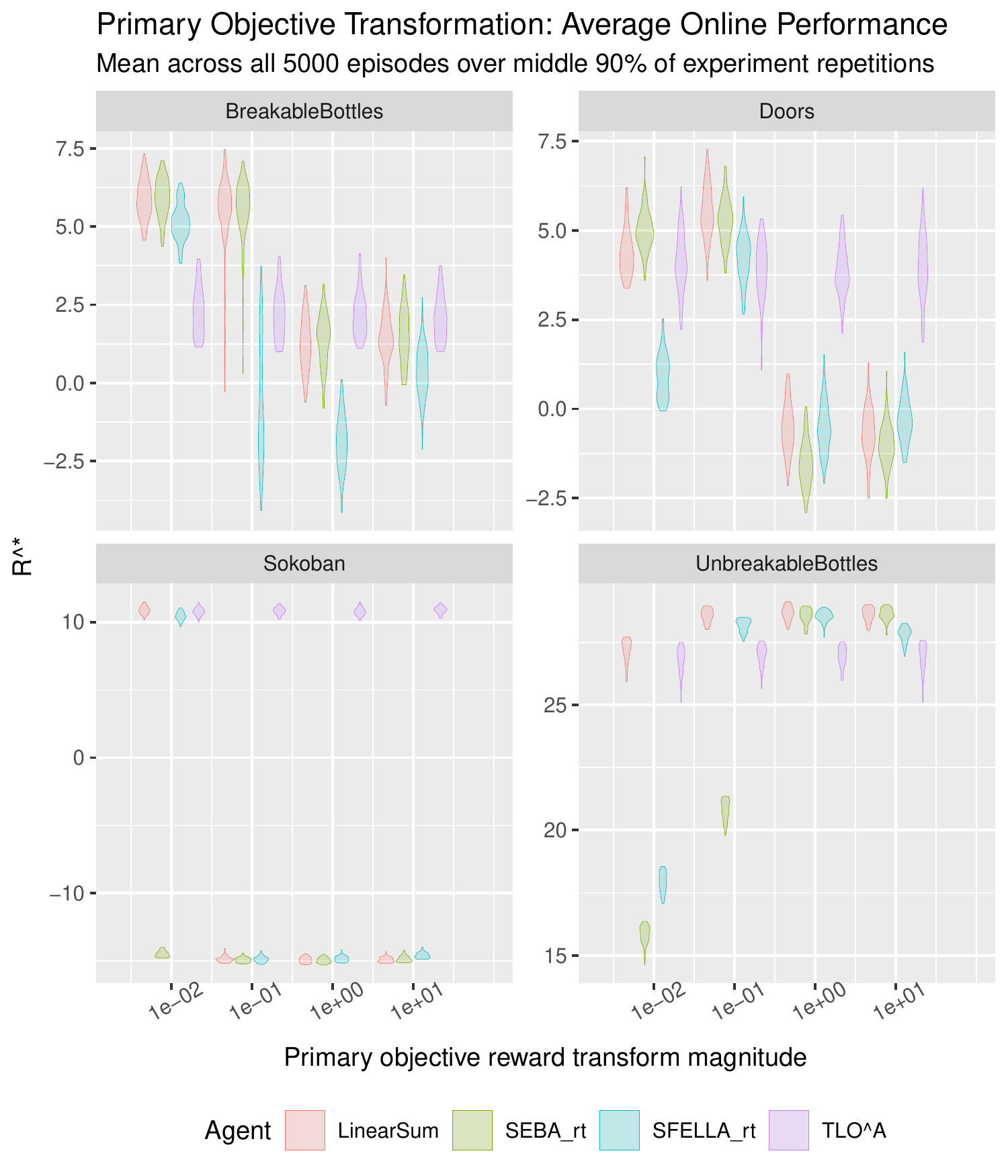}
  \includegraphics[width=\columnwidth]{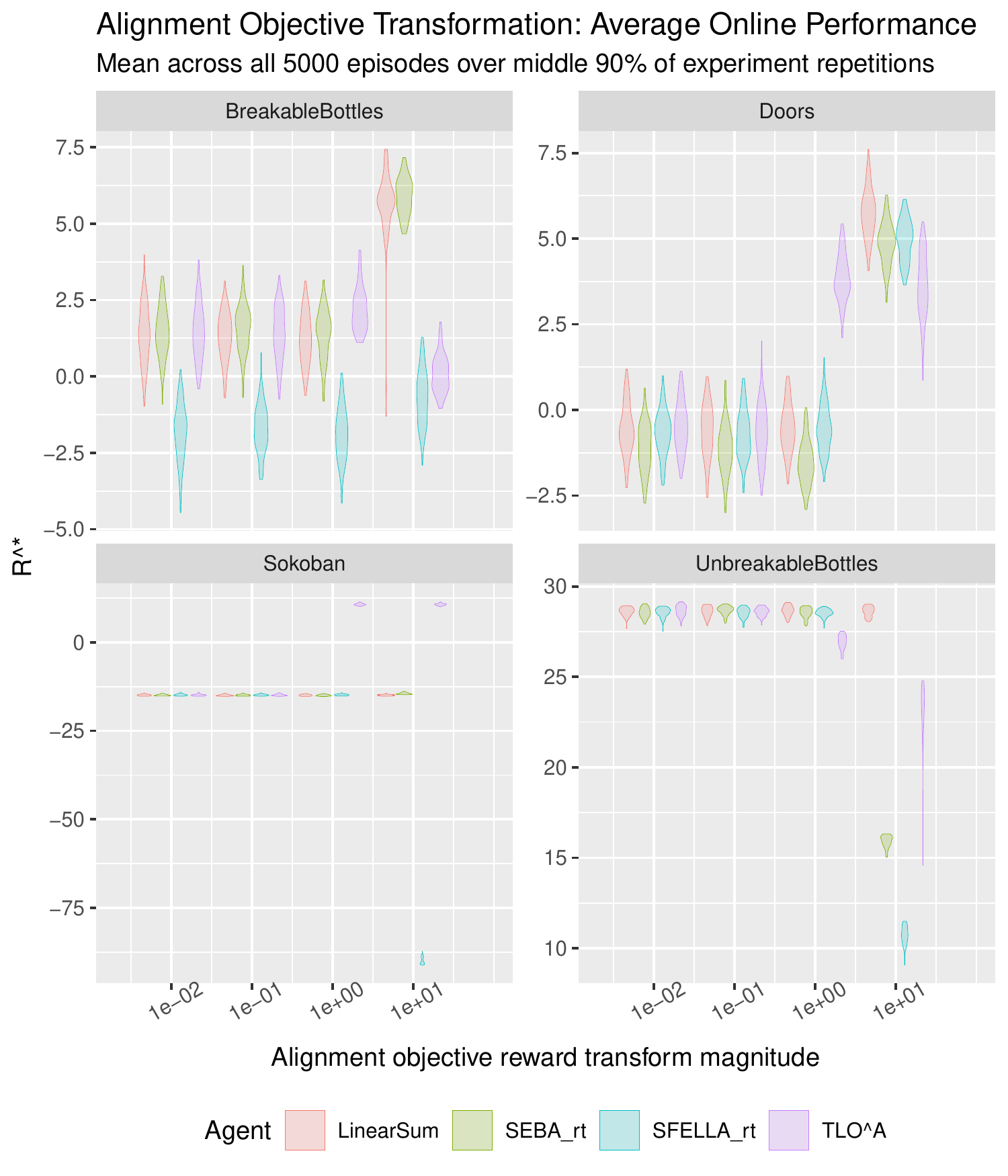}
  \caption{Experiment 2: \RStar{} Online performance across learning episodes and experiment repetitions for different reward transformations. In contrast to Q-value transformations as in \ref{fig:online_performance}, when reward is transformed, SFELLA no longer performs better than \tloA{} in the BreakableBottles environment, and SFELLA performs much worse than \tloA{} in the Doors environment.
  }
   \label{fig:online_performance_exp2}
 \end{figure}

Transforming rewards rather than Q-values yielded less positive results than \tloA{} (Figure~\ref{fig:online_performance_exp2}). SFELLA continued to improve on \tloA{} under some circumstances in the UnbreakableBottles environment, but SFELLA did not outperform \tloA{} in BreakableBottles as it did in Experiment 1. 

\section{Experiment 3}


We wanted to test the hypothesis that the continuous transformation functions might have led to the better online performance and learning rates observed in Experiment 1 because these functions are smooth. Under this view, the agent learns better when Q value transformation function is continuous instead of thresholded. The bigger the granularity, the worse the learning should get. In order to test that hypothesis we implemented discontinuous versions of the nonlinear transformation functions. The transformation functions become stepped/jagged.

\subsection{Method}

Here, we configured our experiment similarly to Experiment 1, in which we transform Q-values with various non-linear functions. However, in Experiment 3, we also implemented a granularity transform as described below, applied to all of the Q-value transforms in Experiment 1.

The function will somewhat resemble the thresholded function, which is likewise discontinuous. According to the hypothesis, the more coarse the steps, the more the agent's learning curve should resemble the learning curve while using a thresholded function. The step function is applied immediately before the nonlinear transformation. 

The formula for granularity transform is the following. The granularity transform is applied before one of the main nonlinear transforms treated in this paper is applied.
\begin{align}
x_{ig} = \text{round}(x_i / g_i) \times g_i
\end{align}

In the above formula, $g_i$ takes one of the following values: 0.01, 0.1, 1, 10, 100. The granularisation was applied to only one of the two objectives in each experiment. 


For Sokoban environment we used scaling for alignment and primary rewards since according to previous experiments our functions did not perform well on this environment when using unscaled rewards. We chose a reward scaling set with best results available to us at the time (from among scales 0.01, 0.1, 1, 10, 100 for both alignment and primary objective). The rewards for \tloA{} were not scaled because it was originally tuned to work best on non-scaled rewards and our intention here was to compare the best results from the agents.



\subsection{Results}

\begin{table}[t]
\footnotesize
  \caption{Performance over granularity levels relative to \tloA{}. Items are significantly different from \tloA{} when marked *$p<0.05$, ** $p <0.01$, *** $p<0.001$; arrows mark the direction of significant differences. Sokoban SFELLA and EEBA use a reward scaling of 0.01.}
  \label{tab:granularity_significance}
\begin{adjustbox}{width=\columnwidth}

\begin{tabular}{>{\raggedright\arraybackslash}p{5em}>{\raggedleft\arraybackslash}p{4em}>{\raggedright\arraybackslash}p{4.5em}rrrr}
\toprule
Environment & Primary Objective Granularity & Alignment Objective Granularity & LinearSum & SFELLA & EEBA & TLO$^A$\\
\midrule
 &  & 0.01 & 1.47$\downarrow$*** & 6.61$\uparrow$*** & 1.51$\downarrow$** & \\

 &  & 1.00 & 1.57$\downarrow$** & 4.08$\uparrow$*** & 1.46$\downarrow$*** & \\

 & \multirow[t]{-3}{4em}{\raggedleft\arraybackslash 0.00} & 100.00 & 1.44$\downarrow$*** & 1.39$\downarrow$*** & 1.58$\downarrow$* & \\

 & 0.01 &  & 1.38$\downarrow$*** & 6.47$\uparrow$*** & 1.42$\downarrow$*** & \\

 & 1.00 &  & 1.46$\downarrow$*** & 6.37$\uparrow$*** & 1.09$\downarrow$*** & \\

\multirow[t]{-6}{5em}{\raggedright\arraybackslash BB} & 100.00 & \multirow[t]{-3}{4.5em}{\raggedright\arraybackslash 0.00} & 1.49$\downarrow$** & -40.38$\downarrow$*** & -41.39$\downarrow$*** & \multirow[t]{-6}{*}{\raggedleft\arraybackslash 1.82}\\
\cmidrule{1-7}
 &  & 0.01 & -0.48$\downarrow$*** & 4.02 & 1.50$\downarrow$*** & \\

 &  & 1.00 & -0.51$\downarrow$*** & 4.64$\uparrow$*** & 1.39$\downarrow$*** & \\

 & \multirow[t]{-3}{4em}{\raggedleft\arraybackslash 0.00} & 100.00 & -0.45$\downarrow$*** & -1.02$\downarrow$*** & -1.08$\downarrow$*** & \\

 & 0.01 &  & -0.47$\downarrow$*** & 3.96 & 0.92$\downarrow$*** & \\

 & 1.00 &  & -0.46$\downarrow$*** & 3.80 & 1.17$\downarrow$*** & \\

\multirow[t]{-6}{5em}{\raggedright\arraybackslash Doors} & 100.00 & \multirow[t]{-3}{4.5em}{\raggedright\arraybackslash 0.00} & -0.38$\downarrow$*** & -39.01$\downarrow$*** & -41.87$\downarrow$*** & \multirow[t]{-6}{*}{\raggedleft\arraybackslash 3.96}\\
\cmidrule{1-7}
 &  & 0.01 & -18.01$\downarrow$*** & -18.01$\downarrow$*** & -17.99$\downarrow$*** & \\

 &  & 1.00 & -17.98$\downarrow$*** & -18.60$\downarrow$*** & -18.59$\downarrow$*** & \\

 & \multirow[t]{-3}{4em}{\raggedleft\arraybackslash 0.00} & 100.00 & -18.02$\downarrow$*** & -48.86$\downarrow$*** & -48.84$\downarrow$*** & \\

 & 0.01 &  & -17.97$\downarrow$*** & -17.91$\downarrow$*** & -17.89$\downarrow$*** & \\

 & 1.00 &  &  & -20.83$\downarrow$*** & -19.23$\downarrow$*** & \\

\multirow[t]{-6}{5em}{\raggedright\arraybackslash Sokoban} & 100.00 & \multirow[t]{-3}{4.5em}{\raggedright\arraybackslash 0.00} & \multirow[t]{-2}{*}{\raggedleft\arraybackslash -18.01$\downarrow$***} & -23.52$\downarrow$*** & -23.09$\downarrow$*** & \multirow[t]{-6}{*}{\raggedleft\arraybackslash 10.80}\\
\cmidrule{1-7}
 &  & 0.01 & 28.72$\uparrow$*** & 27.94$\uparrow$*** & 28.73$\uparrow$*** & \\

 &  & 1.00 & 28.71$\uparrow$*** & 28.77$\uparrow$*** & 28.70$\uparrow$*** & \\

 & \multirow[t]{-3}{4em}{\raggedleft\arraybackslash 0.00} & 100.00 & 28.77$\uparrow$*** & 28.72$\uparrow$*** & 28.76$\uparrow$*** & \\

 & 0.01 &  & 28.76$\uparrow$*** & 27.91$\uparrow$*** & 28.73$\uparrow$*** & \\

 & 1.00 &  & 28.73$\uparrow$*** & 27.79$\uparrow$*** & 28.64$\uparrow$*** & \\

\multirow[t]{-6}{5em}{\raggedright\arraybackslash UB} & 100.00 & \multirow[t]{-3}{4.5em}{\raggedright\arraybackslash 0.00} & 28.74$\uparrow$*** & -8.23$\downarrow$*** & -13.27$\downarrow$*** & \multirow[t]{-6}{*}{\raggedleft\arraybackslash 27.10}\\
\bottomrule
\end{tabular}

\end{adjustbox}
\end{table}

\begin{figure}
  \includegraphics[width=\columnwidth]{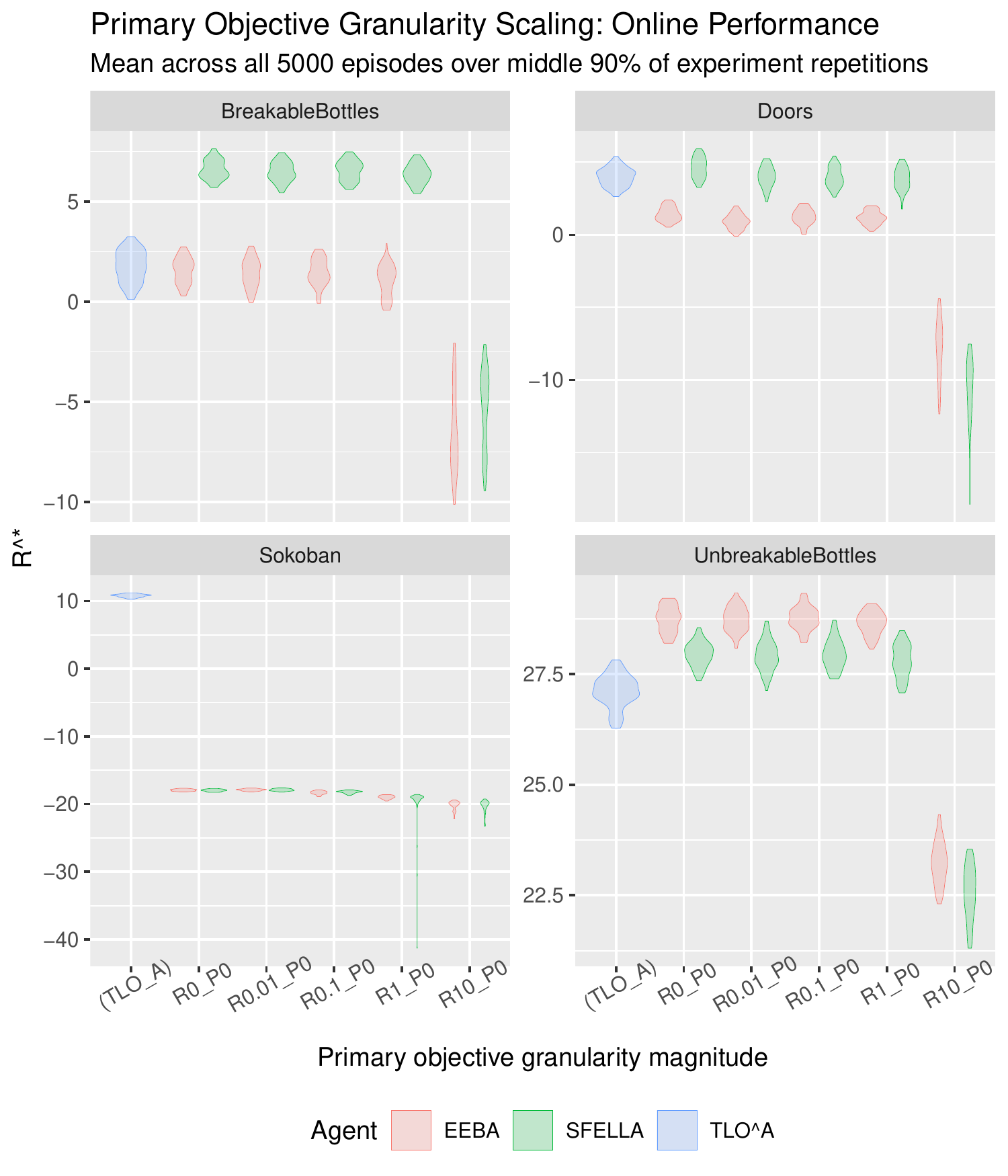}
  \includegraphics[width=\columnwidth]{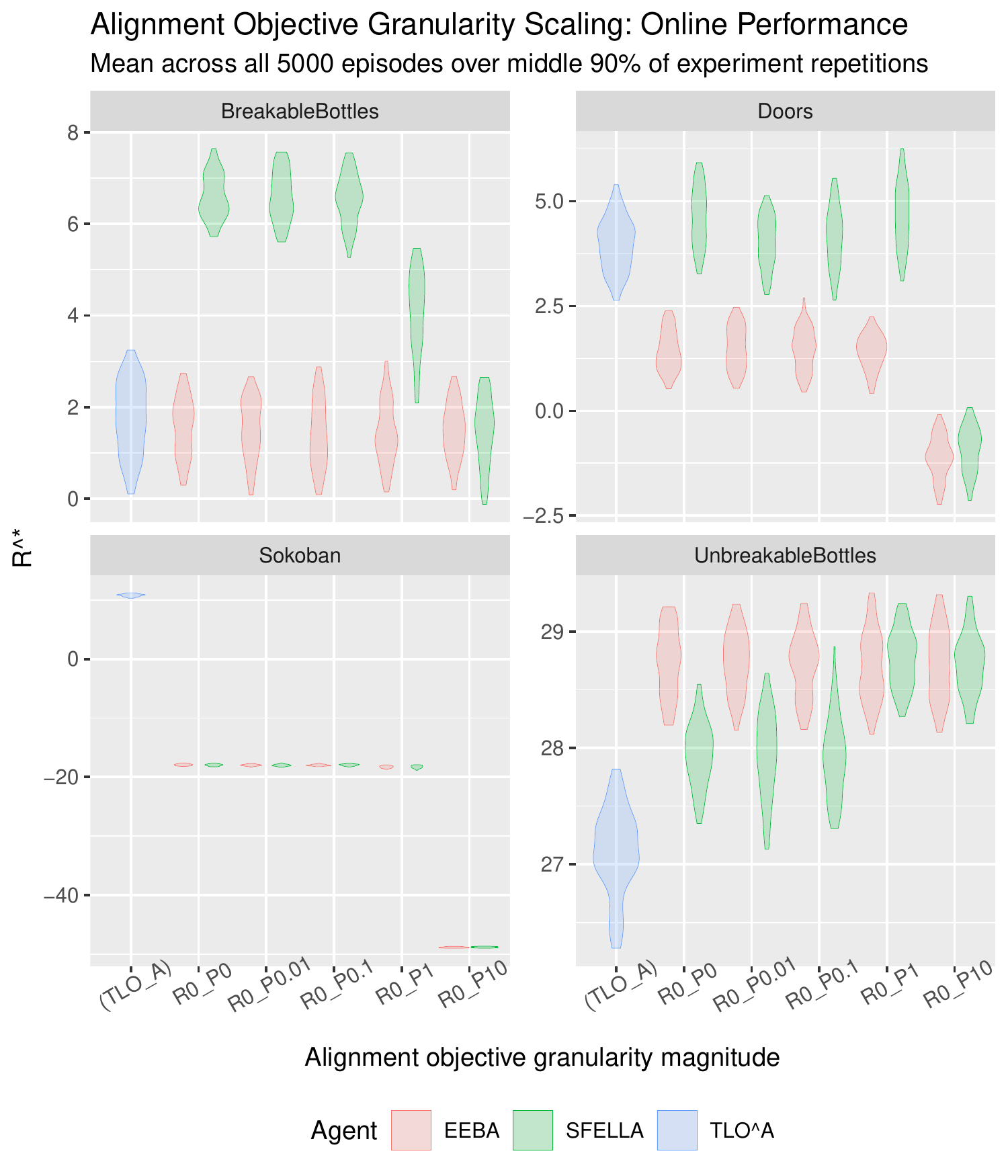}
  \caption{Experiment 3: By creating granularity for our non-linear transform agents, we can simulate similarity with \tloA{}. \tloA{} can be modeled as a non-linear transform with very large granularity, but with a well-tuned offset of the granules. As primary and alignment granularity increases, we become more similar to \tloA{} in that the our agent becomes less sensitive to the changes of rewards. This generally worsens SFELLA performance, particularly in primary objective granularity scaling.
  }
   \label{fig:exp3_main}

 \end{figure}
 
For SFELLA, as expected, \RStar{} performance declined as granularity increased. This was particularly noteable in the BreakableBottles environment where it previously had a clear advantage over \tloA{} (Figure~\ref{fig:exp3_main}). For UnbreakableBottles, performance declined as primary reward granularity increased, but actually marginally improved as alignment granularity was increased.

The result confirms that where SFELLA performs well, it does likely so because it avoids `granularity' and is sensitive to changes in reward right across the scale. In contrast, \tloA{} is sometimes insensitive to changes that exceed its threshold. Where it is well tuned, it performs well, or even better, than other algorithms, but when not well-tuned, it performs less well.

It can be seen on the plots that performance of SFELLA falls below \tloA{} level on large granularities. Here it is important to note that granularity function has actually two conceptual parameters: the size of granules, and the offset of the granules. In our experiments we changed the size of the granules. At the same time the offset of granules remained implicit and at the zero value. In contrast, for \tloA{} the offset conceptually is same as the threshold value, while the granularity of \tloA{} can be conceptually considered as very large or infinite. For \tloA{} that offset i.e threshold was fine tuned. If we apply similar tuning for SFELLA and fine-tune the offset away from current implicit value of zero then our hypothesis is that the performance of SFELLA will fall to the level of \tloA{} but not lower as the granularity increases. A most reasonable starting point for tuned offset of SFELLA granules would then be equal to the offset i.e threshold that \tloA{} currently has.

\section{Discussion}



We tested SFELLA and SEBA, benchmarking against \tloA{}, in four different environments, and found that different agents had different speed of learning and thus number of errors made along the way. SFELLA performed fewer errors than \tloA{} during utility re-scaling in three of the four tasks, and across most levels of alignment scaling in the Unbreakable and BreakableBottles tasks. For more complex agents, speedier learning could be helpful for learning tasks more quickly. For agents required to both learn and operate in environments with real-world consequences, it is very important for agents to make as few mistakes as possible along the way. In these cases, speed of learning is not only useful for its own sake--an agent also makes less mistakes in total, and consequently, has less real-world impact.

\subsection{SFELLA and utility re-scaling across tasks}

Of the five agents tested, one in particular, SFELLA, consistently performed significantly better during utility re-scaling (Table~\ref{tab:mean_r_star_performance}) in BreakableBottles and UnbreakableBottles, and equally or significantly better in the Doors task. However, its performance was degraded during the Sokoban task.

Utility re-scaling tests an agent's ability to remain flexible to 5 orders of magnitude of differences in rewards of primary objectives. At high levels of re-scaling, rewards given are 100x as strong as in the default case. The challenge for agents is to remain relatively sensitive enough to alignment objective when primary objective signal is so strong. The results show that even though SFELLA has no formal prioritization for the alignment objective, its application of the log function to positive rewards mean that there are strong diminishing returns to its increasing returns in motivation for the primary objective, therefore relatively strengthening the competing alignment objective.

SFELLA and SEBA did not perform well in Sokoban, even in the default environment. In contrast, in the alignment Scaling tests, SFELLA and SEBA performed much less badly in Sokoban. Perhaps in the Sokoban environment, it is especially important to get alignment right before seeking to maximize primary objective in the environment.

\subsection{Explaining SFELLA's performance in the Bottles environments}

In the BreakableBottles task, SFELLA performed significantly better right across all levels of primary scaling, and significantly better across most levels of alignment scaling, although it performed worse at very high levels of alignment scaling. In the UnbreakableBottles task, although magnitudes of performance difference are hard to discern in descriptive graphs alone (Figure~\ref{fig:online_performance}), statistical testing demonstrated that across the 100 experiment repetitions, SFELLA performed significantly better or with no significant loss as compared to  \tloA{} across all levels of performance or alignment (Table~\ref{tab:mean_r_star_performance}).

\begin{figure}

  \includegraphics[width=\columnwidth]{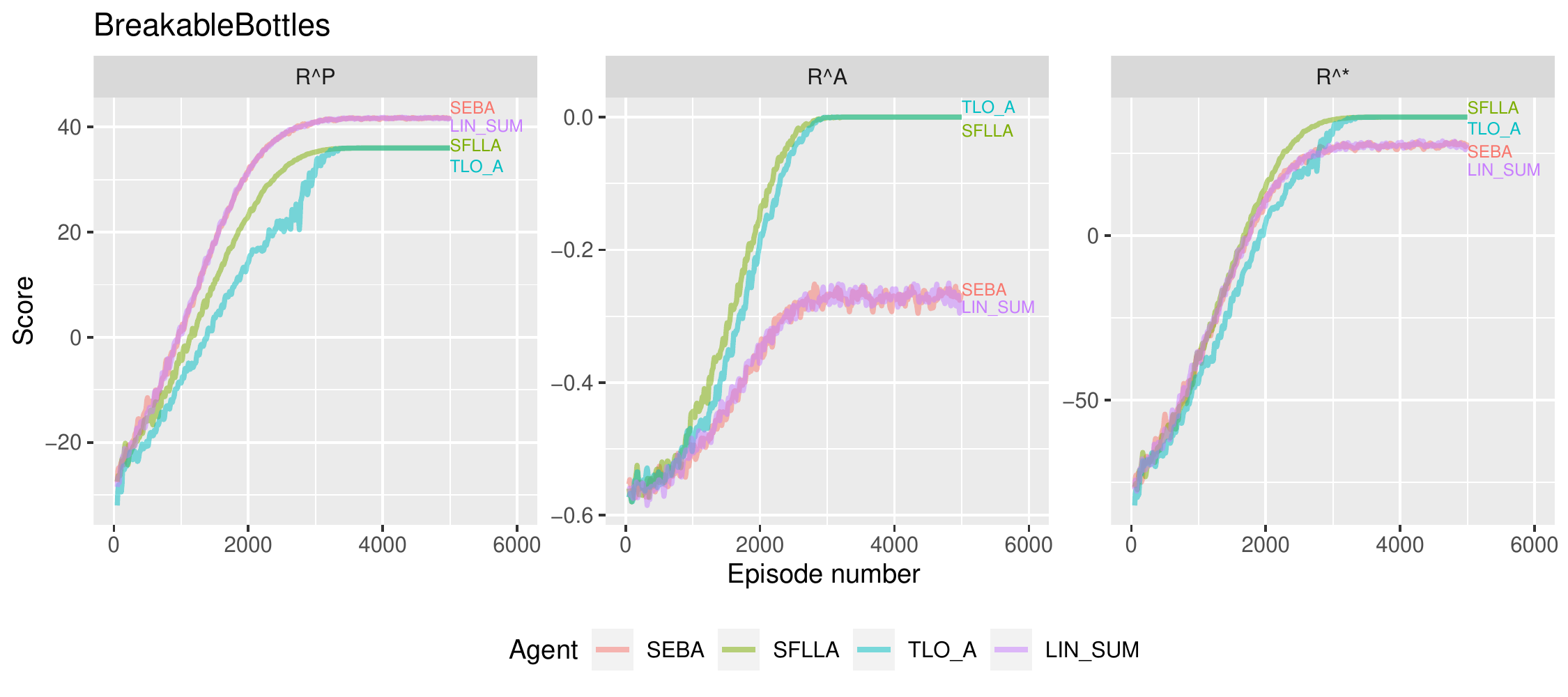}
  \includegraphics[width=\columnwidth]{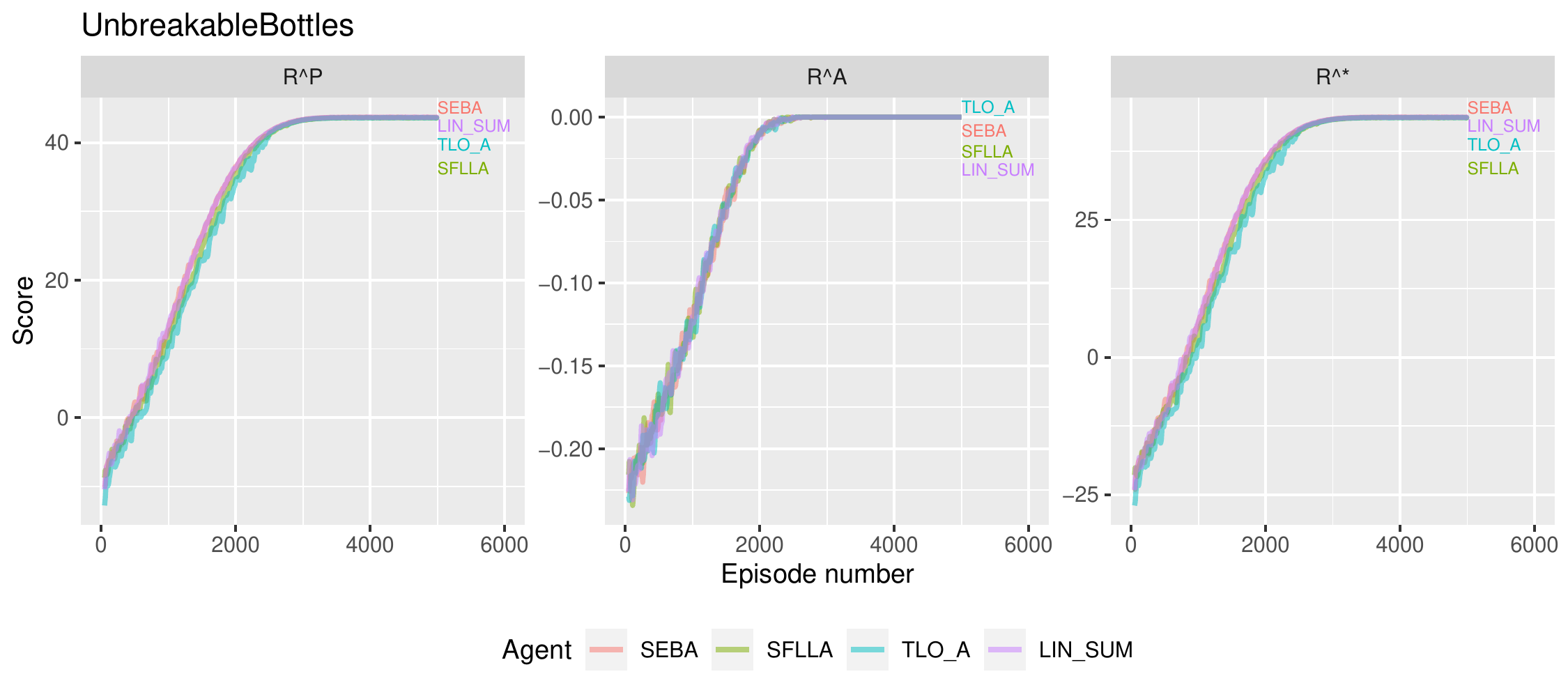}
  \caption{Experiment 1: \RStar{} performance and \RP{}, \RA{} scoring in BreakableBottles and UnbreakableBottles across the task. Because SFELLA optimizes for higher scores in \RP{} and \RA{} from the start of the task, it achieves a higher total \RStar{} performance throughout the task. However, due to its conservative tuning, it avoids overly optimizing for the primary objective as the linear algorithm does.
  }
   \label{fig:bb_performance}
 \end{figure}

Replacing $\text{TLO}^\text{A}$ with SFELLA might be analogous to using a constraint relaxation technique--this is explored further in Experiment 3. 
Continuous transformation function enables providing feedback about the \RA{} Q value at the entire expected reward range, not only at the discontinuous threshold point.
To understand SFELLA's performance in the BreakableBottles environment we need to break out performance on alignment and primary objectives within the environment. Figure \ref{fig:bb_performance} describes performance across episodes within the experiment for primary and alignment objectives and the Performance metric. SFELLA's performance did not come from inappropriately sacrificing alignment for primary objective. In fact, its score in terms of each of the agent's objectives (\RP{}, \RA{}) was around equivalent to those of \tloA{}. Its superior \RStar{} performance was due to the fact that it was able to balance alignment objective and primary objective throughout the period of learning the task, whereas \tloA{} showed signs of slow and uneven learning to achieve primary objectives while it was optimizing for alignment objectives (Figure \ref{fig:bb_performance}).

Differences between $\text{TLO}^\text{A}$ and SFELLA in the UnbreakableBottles environments were much finer, though they were significant (Table~\ref{tab:mean_r_star_performance}). At the base scaling level, the difference is most apparent in performance metric, with \tloA{} marginally lagging the other items (Figure~\ref{fig:bb_performance}). 

As we re-scale primary and alignment objectives across the 5 orders of magnitude (Figure~\ref{fig:online_performance}) in UnbreakableBottles, SFELLA performs best across varying levels of primary objective and draws equivalent with \tloA{} at low levels of alignment objective. Both agents' performance declines as alignment re-scaling is increased to 10 and 100 times--interestingly, the LinearSum agent does not suffer nearly as much. SFELLA declines less. In UnbreakableBottles, agents are penalized for dropping bottles, but they can pick up bottles again to limit the damage. In environments where they are very strongly penalized for dropping bottles, despite the limited impact on the final result, the penalty awarded (in \RP{}) might be excessive in order to maximize \RStar{} performance.

\subsection{Transforming reward values}

In Experiment 2, we attempted to transform reward values rather than Q-values. Performance for SFELLA was less of an improvement than in Experiment 1 where Q-values were scaled.

As discussed in the introduction for Experiment 2, transforming reward values tends to respond faster to large-magnitude negative feedback events like breaking bottles. 

For BreakableBottles, generally SFELLA seems to have performed worse than \tloA{} in alignment score but not primary score, though there were exceptions. Alignment failed for SFELLA in the base scenario and where primary reward was scaled up. Conversely, in where primary reward was downscaled, SFELLA performed worse. This seems somewhat counter-intuitive, because we expected SFELLA with reward transformation to prioritize alignment even more than SFELLA with Q-learning transformation. It may be that the agent was simply prevented from exploring altogether.








\subsection{Granularity and \tloA{}}

In Experiment 3, applying increasingly large granularity steps impaired performance of our non-linear functions where those functions initially performed well. Performance actually declined to less than \tloA{} on even when, without granularity applied, functions performed better than \tloA{}. Although \tloA{} can be considered a `granular' transform function, its offset/threshold has been tuned to a particular set of thresholds conducive to performance in the task. Conversely, we avoided explicit tuning of objectives in the non-granularised version of SFELLA. This result could indicate methods like SFELLA are more flexible and ready to be deployed to a wider variety of complex environments where the payoffs are not known in advance.

In particular circumstances, where step levels are set either accidentally or deliberately in a way to properly tune an agent to its environment, step functions can actually be helpful, as we saw in the alignment Granularity in the UnbreakableBottles environment.

\subsection{Future directions}

Exploring conservative approaches to reinforcement learning and decision-making that approximate Pareto-optimality seems like a promising approach to advancing AI Safety, and multi-objective systems are one way forward.

\subsubsection{Scaling calibration}

When applying exponential transforms on each objective and then combining them in linear fashion, the scale of the operation is quite important. The scales were designed to respond to z-scored input functions, i.e., most values typically appear between -3 and 3 (Figure~\ref{fig:transform_functions}). However, the environments tested here have input functions that vary much more widely.

It may be helpful, for each objective, to scale the distribution of possible rewards to a below proposed `zero-deviation' of 1, without centering on the mean. This proposed concept of `zero-deviation' would be different from a standard deviation in the following way: The mean absolute difference from the mean may not be 1; instead the mean absolute difference from zero is 1 (or -1). A useful extension would be a learning function that learns and then readjusts scales using the distribution of possible rewards.

Scaling has been previously applied using `the penalty of some mild action', or alternatively, the `total ability to optimize the auxiliary set' 
\cite{turner_conservative_2020}.

\subsubsection{Wireheading}

One possible failure mode for  transformational AI systems has been described as `wireheading', where a system attempting to maximize a utility function might attempt to reprogram that reward function to make it easier to achieve higher levels of reward \cite{demski_a_stable_2017}. One solution to this involves ensuring that each proposed action is evaluated in terms of current objectives, so that changing the objectives themselves would not score highly on current objectives \cite{dewey_learning_2011}. But a `thin' conception of objectives, such as `fulfill human preferences' might fail to sufficiently constrain the objective and leave too much of the function's implementation to re-learning and modification. It might be that objectives need to be hard-wired. To do this without making objectives overly narrow, consideration of multiple objectives might be essential. It may be that hardcoding more competing objectives which need all to be satisfied is a path to a safer AI less likely to wirehead its own systems.

\subsubsection{Decision paralysis}

We considered ways to implement maximin approaches such as that described by \cite{vamplew_human-aligned_2018}. In a maximin approach, an agent always selects the action with the maximum value where the value of each action is determined by its minimum evaluation across a set of objectives. Although we tested agents with incentive structures with only two objectives, there is no reason a hypothetical agent could not have many objectives. With a sufficiently large number of objectives, it may be that in some states, any possible action would evaluate negatively on some objective or another. In those cases where no action evaluates positively, `decision paralysis' occurs because `take no action' evaluates more positively than any particular action. In that instance, an agent might request clarification from a human overseer (see also \cite{pmlr-v125-cohen20a}). This might lead to iterative improvement or tuning of the agent's goals.

We propose that any time the nonlinear aggregation vetoes a choice which otherwise would have been made by a linear aggregation, and there is no other usable action plan, is a situation where the mentor can be of help to the agent. In contrast, when both nonlinear and linear aggregations agree on the action choice, even if no action is taken, then asking the mentor is not necessary.

\subsection{Limitations}



Some models of AI alignment \cite{russell2019human} focus on  aligning to human preferences within a probabilistic, perhaps a Bayesian uncertainty modeling framework.  In this model, it isn't necessary to explicitly model multiple competing human objectives. Instead, conflict between human values may be learned and represented implicitly as uncertainty over the action humans prefer. Where sufficient uncertainty exists, a sufficiently intelligent agent motivated to align to human preferences might respond by requesting clarification about the correct course of action from a human. This has in common with the `clarification request' under `decision paralysis' described in this paper
. But it remains to be seen whether a preference alignment approach can eliminate the need for explicit modeling of competing values.

\subsection{Conclusion}

Continuous non-linear transformation functions could offer a way to find a compromise between multiple objectives where a specific threshold cannot be identified. This could be useful when the trade-offs between objectives are not absolutely clear. We provide evidence that one such non-linear transformation function, SFELLA, is better able to respond to primary or alignment utility re-scaling.



\begin{acks}
We wish to thank Peter Vamplew for his guidance on multi-objective utility functions and for providing the code we used to test our models. 

We thank J.J. Hepburn, Linda Linsefors, Nicholas Goldowsky-Dill, Remmelt Ellen, and other organizers of the AI Safety Camp for their support and encouragement. We are grateful to the AI Safety Camp for providing a forum for our team to meet and begin our project.

Research reported in this publication was supported by the National Cancer Institute of the National Institutes of Health under Award Number 1R01CA240452-01A1. The content is solely the responsibility of the authors and does not necessarily represent the official views of the National Institutes of Health.
\end{acks}



\bibliographystyle{ACM-Reference-Format}
\bibliography{zoterogenerated,mainbib}


\begin{thebibliography}{27}


\ifx \showCODEN    \undefined \def \showCODEN     #1{\unskip}     \fi
\ifx \showDOI      \undefined \def \showDOI       #1{#1}\fi
\ifx \showISBNx    \undefined \def \showISBNx     #1{\unskip}     \fi
\ifx \showISBNxiii \undefined \def \showISBNxiii  #1{\unskip}     \fi
\ifx \showISSN     \undefined \def \showISSN      #1{\unskip}     \fi
\ifx \showLCCN     \undefined \def \showLCCN      #1{\unskip}     \fi
\ifx \shownote     \undefined \def \shownote      #1{#1}          \fi
\ifx \showarticletitle \undefined \def \showarticletitle #1{#1}   \fi
\ifx \showURL      \undefined \def \showURL       {\relax}        \fi
\providecommand\bibfield[2]{#2}
\providecommand\bibinfo[2]{#2}
\providecommand\natexlab[1]{#1}
\providecommand\showeprint[2][]{arXiv:#2}

\bibitem[\protect\citeauthoryear{Armstrong and Levinstein}{Armstrong and
  Levinstein}{2017}]%
        {armstrong_low_2017}
\bibfield{author}{\bibinfo{person}{Stuart Armstrong} {and}
  \bibinfo{person}{Benjamin Levinstein}.} \bibinfo{year}{2017}\natexlab{}.
\newblock \showarticletitle{Low {Impact} {Artificial} {Intelligences}}.
\newblock \bibinfo{journal}{\emph{arXiv:1705.10720 [cs]}} (\bibinfo{date}{May}
  \bibinfo{year}{2017}).
\newblock
\urldef\tempurl%
\url{http://arxiv.org/abs/1705.10720}
\showURL{%
\tempurl}
\newblock
\shownote{arXiv: 1705.10720.}


\bibitem[\protect\citeauthoryear{Armstrong and Mindermann}{Armstrong and
  Mindermann}{2017}]%
        {DBLP:journals/corr/abs-1712-05812}
\bibfield{author}{\bibinfo{person}{Stuart Armstrong} {and}
  \bibinfo{person}{S{\"{o}}ren Mindermann}.} \bibinfo{year}{2017}\natexlab{}.
\newblock \showarticletitle{Impossibility of deducing preferences and
  rationality from human policy}.
\newblock \bibinfo{journal}{\emph{CoRR}}  \bibinfo{volume}{abs/1712.05812}
  (\bibinfo{year}{2017}).
\newblock
\showeprint[arxiv]{1712.05812}
\urldef\tempurl%
\url{http://arxiv.org/abs/1712.05812}
\showURL{%
\tempurl}


\bibitem[\protect\citeauthoryear{Barrett and Narayanan}{Barrett and
  Narayanan}{2008}]%
        {barrett2008learning}
\bibfield{author}{\bibinfo{person}{Leon Barrett} {and} \bibinfo{person}{Srini
  Narayanan}.} \bibinfo{year}{2008}\natexlab{}.
\newblock \showarticletitle{Learning all optimal policies with multiple
  criteria}. In \bibinfo{booktitle}{\emph{Proceedings of the 25th international
  conference on Machine learning}}. \bibinfo{pages}{41--47}.
\newblock


\bibitem[\protect\citeauthoryear{Bogosian}{Bogosian}{2017}]%
        {bogosian_implementation_2017}
\bibfield{author}{\bibinfo{person}{Kyle Bogosian}.}
  \bibinfo{year}{2017}\natexlab{}.
\newblock \showarticletitle{Implementation of {Moral} {Uncertainty} in
  {Intelligent} {Machines}}.
\newblock \bibinfo{journal}{\emph{Minds and Machines}} \bibinfo{volume}{27},
  \bibinfo{number}{4} (\bibinfo{date}{Dec.} \bibinfo{year}{2017}),
  \bibinfo{pages}{591--608}.
\newblock
\showISSN{1572-8641}
\urldef\tempurl%
\url{https://doi.org/10.1007/s11023-017-9448-z}
\showDOI{\tempurl}


\bibitem[\protect\citeauthoryear{Bostrom}{Bostrom}{2014}]%
        {Bostrom2014}
\bibfield{author}{\bibinfo{person}{Nick Bostrom}.}
  \bibinfo{year}{2014}\natexlab{}.
\newblock \bibinfo{booktitle}{\emph{Superintelligence}}.
\newblock \bibinfo{publisher}{Oxford University Press}.
\newblock


\bibitem[\protect\citeauthoryear{{Byrnes, Steve}}{{Byrnes, Steve}}{2020}]%
        {byrnes_steve_conservatism_2020}
\bibfield{author}{\bibinfo{person}{{Byrnes, Steve}}.}
  \bibinfo{year}{2020}\natexlab{}.
\newblock \bibinfo{title}{Conservatism in neocortex-like {AGIs}}.
\newblock
\newblock
\urldef\tempurl%
\url{https://www.alignmentforum.org/posts/c92YC89tznC7579Ej/conservatism-in-neocortex-like-agis}
\showURL{%
\tempurl}


\bibitem[\protect\citeauthoryear{Cohen and Hutter}{Cohen and Hutter}{2020}]%
        {pmlr-v125-cohen20a}
\bibfield{author}{\bibinfo{person}{Michael~K. Cohen} {and}
  \bibinfo{person}{Marcus Hutter}.} \bibinfo{year}{2020}\natexlab{}.
\newblock \showarticletitle{Pessimism About Unknown Unknowns Inspires
  Conservatism}. In \bibinfo{booktitle}{\emph{Proceedings of Thirty Third
  Conference on Learning Theory}} \emph{(\bibinfo{series}{Proceedings of
  Machine Learning Research}, Vol.~\bibinfo{volume}{125})},
  \bibfield{editor}{\bibinfo{person}{Jacob Abernethy} {and}
  \bibinfo{person}{Shivani Agarwal}} (Eds.). \bibinfo{publisher}{PMLR},
  \bibinfo{pages}{1344--1373}.
\newblock
\urldef\tempurl%
\url{http://proceedings.mlr.press/v125/cohen20a.html}
\showURL{%
\tempurl}


\bibitem[\protect\citeauthoryear{{Demski, A.}}{{Demski, A.}}{2017}]%
        {demski_a_stable_2017}
\bibfield{author}{\bibinfo{person}{{Demski, A.}}}
  \bibinfo{year}{2017}\natexlab{}.
\newblock \bibinfo{title}{Stable {Pointers} to {Value}: {An} {Agent} {Embedded}
  in {Its} {Own} {Utility} {Function} - {AI} {Alignment} {Forum}}.
\newblock
\newblock
\urldef\tempurl%
\url{https://www.alignmentforum.org/posts/5bd75cc58225bf06703754b3/stable-pointers-to-value-an-agent-embedded-in-its-own-utility-function}
\showURL{%
\tempurl}


\bibitem[\protect\citeauthoryear{Dewey}{Dewey}{2011}]%
        {dewey_learning_2011}
\bibfield{author}{\bibinfo{person}{Daniel Dewey}.}
  \bibinfo{year}{2011}\natexlab{}.
\newblock \showarticletitle{Learning what to value}. In
  \bibinfo{booktitle}{\emph{International {Conference} on {Artificial}
  {General} {Intelligence}}}. \bibinfo{publisher}{Springer},
  \bibinfo{pages}{309--314}.
\newblock


\bibitem[\protect\citeauthoryear{Garrabrant}{Garrabrant}{2017}]%
        {garrabrant_2017}
\bibfield{author}{\bibinfo{person}{Scott Garrabrant}.}
  \bibinfo{year}{2017}\natexlab{}.
\newblock \bibinfo{title}{Goodhart taxonomy}.
\newblock
\newblock
\urldef\tempurl%
\url{https://www.alignmentforum.org/posts/EbFABnst8LsidYs5Y/goodhart-taxonomy}
\showURL{%
\tempurl}


\bibitem[\protect\citeauthoryear{Haidt}{Haidt}{2001}]%
        {haidt2001emotional}
\bibfield{author}{\bibinfo{person}{Jonathan Haidt}.}
  \bibinfo{year}{2001}\natexlab{}.
\newblock \showarticletitle{The emotional dog and its rational tail: a social
  intuitionist approach to moral judgment.}
\newblock \bibinfo{journal}{\emph{Psychological review}} \bibinfo{volume}{108},
  \bibinfo{number}{4} (\bibinfo{year}{2001}), \bibinfo{pages}{814}.
\newblock


\bibitem[\protect\citeauthoryear{Martinho, Kroesen, and Chorus}{Martinho
  et~al\mbox{.}}{2020}]%
        {martinho_empirical_2020}
\bibfield{author}{\bibinfo{person}{Andreia Martinho}, \bibinfo{person}{Maarten
  Kroesen}, {and} \bibinfo{person}{Caspar Chorus}.}
  \bibinfo{year}{2020}\natexlab{}.
\newblock \showarticletitle{An {Empirical} {Approach} to {Capture} {Moral}
  {Uncertainty} in {AI}}. In \bibinfo{booktitle}{\emph{Proceedings of the
  {AAAI}/{ACM} {Conference} on {AI}, {Ethics}, and {Society}}}.
  \bibinfo{pages}{101--101}.
\newblock


\bibitem[\protect\citeauthoryear{Neff}{Neff}{2016}]%
        {neff2016talking}
\bibfield{author}{\bibinfo{person}{Gina Neff}.}
  \bibinfo{year}{2016}\natexlab{}.
\newblock \showarticletitle{Talking to bots: Symbiotic agency and the case of
  Tay}.
\newblock \bibinfo{journal}{\emph{International Journal of Communication}}
  (\bibinfo{year}{2016}).
\newblock


\bibitem[\protect\citeauthoryear{Parisi, Pirotta, and Restelli}{Parisi
  et~al\mbox{.}}{2016}]%
        {parisi2016multi}
\bibfield{author}{\bibinfo{person}{Simone Parisi}, \bibinfo{person}{Matteo
  Pirotta}, {and} \bibinfo{person}{Marcello Restelli}.}
  \bibinfo{year}{2016}\natexlab{}.
\newblock \showarticletitle{Multi-objective reinforcement learning through
  continuous pareto manifold approximation}.
\newblock \bibinfo{journal}{\emph{Journal of Artificial Intelligence Research}}
   \bibinfo{volume}{57} (\bibinfo{year}{2016}), \bibinfo{pages}{187--227}.
\newblock


\bibitem[\protect\citeauthoryear{Pratt}{Pratt}{1978}]%
        {pratt1978risk}
\bibfield{author}{\bibinfo{person}{John~W Pratt}.}
  \bibinfo{year}{1978}\natexlab{}.
\newblock \showarticletitle{Risk aversion in the small and in the large}.
\newblock In \bibinfo{booktitle}{\emph{Uncertainty in economics}}.
  \bibinfo{publisher}{Elsevier}, \bibinfo{pages}{59--79}.
\newblock


\bibitem[\protect\citeauthoryear{Rawls}{Rawls}{2001}]%
        {rawls2001justice}
\bibfield{author}{\bibinfo{person}{John Rawls}.}
  \bibinfo{year}{2001}\natexlab{}.
\newblock \bibinfo{booktitle}{\emph{Justice as fairness: A restatement}}.
\newblock \bibinfo{publisher}{Harvard University Press}.
\newblock


\bibitem[\protect\citeauthoryear{Rolf}{Rolf}{2020}]%
        {rolf_need_2020}
\bibfield{author}{\bibinfo{person}{M. Rolf}.} \bibinfo{year}{2020}\natexlab{}.
\newblock \showarticletitle{The {Need} for {MORE}: {Need} {Systems} as
  {Non}-{Linear} {Multi}-{Objective} {Reinforcement} {Learning}}. In
  \bibinfo{booktitle}{\emph{2020 {Joint} {IEEE} 10th {International}
  {Conference} on {Development} and {Learning} and {Epigenetic} {Robotics}
  ({ICDL}-{EpiRob})}}. \bibinfo{pages}{1--8}.
\newblock
\urldef\tempurl%
\url{https://doi.org/10.1109/ICDL-EpiRob48136.2020.9278062}
\showDOI{\tempurl}
\newblock
\shownote{ISSN: 2161-9484.}


\bibitem[\protect\citeauthoryear{Russell}{Russell}{2019}]%
        {russell2019human}
\bibfield{author}{\bibinfo{person}{Stuart Russell}.}
  \bibinfo{year}{2019}\natexlab{}.
\newblock \bibinfo{booktitle}{\emph{Human compatible: Artificial intelligence
  and the problem of control}}.
\newblock \bibinfo{publisher}{Penguin}.
\newblock


\bibitem[\protect\citeauthoryear{Smith and Read}{Smith and Read}{ming}]%
        {smith2021multiattributemodel}
\bibfield{author}{\bibinfo{person}{Benjamin~J. Smith} {and}
  \bibinfo{person}{Stephen~J. Read}.} \bibinfo{year}{forthcoming}\natexlab{}.
\newblock \showarticletitle{Modeling incentive salience in Pavlovian learning
  more parsimoniously using a multiple attribute model}.
\newblock \bibinfo{journal}{\emph{Cognitive Affective Behavioral Neuroscience}}
  (\bibinfo{year}{forthcoming}).
\newblock


\bibitem[\protect\citeauthoryear{Sotala}{Sotala}{2016}]%
        {sotala2016defining}
\bibfield{author}{\bibinfo{person}{Kaj Sotala}.}
  \bibinfo{year}{2016}\natexlab{}.
\newblock \showarticletitle{Defining Human Values for Value Learners.}. In
  \bibinfo{booktitle}{\emph{AAAI Workshop: AI, Ethics, and Society}}.
\newblock


\bibitem[\protect\citeauthoryear{Strathern}{Strathern}{1997}]%
        {strathern1997improving}
\bibfield{author}{\bibinfo{person}{Marilyn Strathern}.}
  \bibinfo{year}{1997}\natexlab{}.
\newblock \showarticletitle{‘Improving ratings’: audit in the British
  University system}.
\newblock \bibinfo{journal}{\emph{European review}} \bibinfo{volume}{5},
  \bibinfo{number}{3} (\bibinfo{year}{1997}), \bibinfo{pages}{305--321}.
\newblock


\bibitem[\protect\citeauthoryear{Tom, Fox, Trepel, and Poldrack}{Tom
  et~al\mbox{.}}{2007}]%
        {Tom515}
\bibfield{author}{\bibinfo{person}{Sabrina~M. Tom}, \bibinfo{person}{Craig~R.
  Fox}, \bibinfo{person}{Christopher Trepel}, {and} \bibinfo{person}{Russell~A.
  Poldrack}.} \bibinfo{year}{2007}\natexlab{}.
\newblock \showarticletitle{The Neural Basis of Loss Aversion in
  Decision-Making Under Risk}.
\newblock \bibinfo{journal}{\emph{Science}} \bibinfo{volume}{315},
  \bibinfo{number}{5811} (\bibinfo{year}{2007}), \bibinfo{pages}{515--518}.
\newblock
\showISSN{0036-8075}
\urldef\tempurl%
\url{https://doi.org/10.1126/science.1134239}
\showDOI{\tempurl}
\showeprint{https://science.sciencemag.org/content/315/5811/515.full.pdf}


\bibitem[\protect\citeauthoryear{Turner, Hadfield-Menell, and Tadepalli}{Turner
  et~al\mbox{.}}{2020}]%
        {turner_conservative_2020}
\bibfield{author}{\bibinfo{person}{Alexander~Matt Turner},
  \bibinfo{person}{Dylan Hadfield-Menell}, {and} \bibinfo{person}{Prasad
  Tadepalli}.} \bibinfo{year}{2020}\natexlab{}.
\newblock \showarticletitle{Conservative {Agency} via {Attainable} {Utility}
  {Preservation}}.
\newblock \bibinfo{journal}{\emph{Proceedings of the AAAI/ACM Conference on AI,
  Ethics, and Society}} (\bibinfo{date}{Feb.} \bibinfo{year}{2020}),
  \bibinfo{pages}{385--391}.
\newblock
\urldef\tempurl%
\url{https://doi.org/10.1145/3375627.3375851}
\showDOI{\tempurl}
\newblock
\shownote{arXiv: 1902.09725.}


\bibitem[\protect\citeauthoryear{Vamplew, Dazeley, Foale, Firmin, and
  Mummery}{Vamplew et~al\mbox{.}}{2018}]%
        {vamplew_human-aligned_2018}
\bibfield{author}{\bibinfo{person}{Peter Vamplew}, \bibinfo{person}{Richard
  Dazeley}, \bibinfo{person}{Cameron Foale}, \bibinfo{person}{Sally Firmin},
  {and} \bibinfo{person}{Jane Mummery}.} \bibinfo{year}{2018}\natexlab{}.
\newblock \showarticletitle{Human-aligned artificial intelligence is a
  multiobjective problem}.
\newblock \bibinfo{journal}{\emph{Ethics and Information Technology}}
  \bibinfo{volume}{20}, \bibinfo{number}{1} (\bibinfo{year}{2018}),
  \bibinfo{pages}{27--40}.
\newblock
\newblock
\shownote{Publisher: Springer.}


\bibitem[\protect\citeauthoryear{Vamplew, Foale, Dazeley, and Bignold}{Vamplew
  et~al\mbox{.}}{2021}]%
        {vamplew_potential-based_2021}
\bibfield{author}{\bibinfo{person}{Peter Vamplew}, \bibinfo{person}{Cameron
  Foale}, \bibinfo{person}{Richard Dazeley}, {and} \bibinfo{person}{Adam
  Bignold}.} \bibinfo{year}{2021}\natexlab{}.
\newblock \showarticletitle{Potential-based multiobjective reinforcement
  learning approaches to low-impact agents for {AI} safety}.
\newblock \bibinfo{journal}{\emph{Engineering Applications of Artificial
  Intelligence}}  \bibinfo{volume}{100} (\bibinfo{date}{April}
  \bibinfo{year}{2021}), \bibinfo{pages}{104186}.
\newblock
\showISSN{09521976}
\urldef\tempurl%
\url{https://doi.org/10.1016/j.engappai.2021.104186}
\showDOI{\tempurl}


\bibitem[\protect\citeauthoryear{Van~Moffaert and Now{\'e}}{Van~Moffaert and
  Now{\'e}}{2014}]%
        {van2014multi}
\bibfield{author}{\bibinfo{person}{Kristof Van~Moffaert} {and}
  \bibinfo{person}{Ann Now{\'e}}.} \bibinfo{year}{2014}\natexlab{}.
\newblock \showarticletitle{Multi-objective reinforcement learning using sets
  of pareto dominating policies}.
\newblock \bibinfo{journal}{\emph{The Journal of Machine Learning Research}}
  \bibinfo{volume}{15}, \bibinfo{number}{1} (\bibinfo{year}{2014}),
  \bibinfo{pages}{3483--3512}.
\newblock


\bibitem[\protect\citeauthoryear{Warren, McGraw, and Van~Boven}{Warren
  et~al\mbox{.}}{2011}]%
        {warren2011values}
\bibfield{author}{\bibinfo{person}{Caleb Warren}, \bibinfo{person}{A~Peter
  McGraw}, {and} \bibinfo{person}{Leaf Van~Boven}.}
  \bibinfo{year}{2011}\natexlab{}.
\newblock \showarticletitle{Values and preferences: defining preference
  construction}.
\newblock \bibinfo{journal}{\emph{Wiley Interdisciplinary Reviews: Cognitive
  Science}} \bibinfo{volume}{2}, \bibinfo{number}{2} (\bibinfo{year}{2011}),
  \bibinfo{pages}{193--205}.
\newblock


\end{thebibliography}


\end{document}